\documentclass[sensors,article,submit,pdftex,moreauthors]{Definitions/mdpi} 

\firstpage{1} 
\pubvolume{1}
\issuenum{1}
\articlenumber{0}
\pubyear{2023}
\copyrightyear{2023}
\datereceived{ } 
\daterevised{ } 
\dateaccepted{ } 
\datepublished{ } 
\hreflink{https://doi.org/} 

\usepackage{subcaption}
\newcommand{\argmax}{\mathop{\mathrm{argmax}}\limits}

\Title{Compensating for Sensing Failures via Delegation in Human-AI Hybrid Systems}

\TitleCitation{Compensating for Sensing Failures via Delegation in Human-AI Hybrid Systems}

\Author{Andrew Fuchs $^{1,2,*}$\orcidA{}, Andrea Passarella $^{2}$\orcidB{} and Marco Conti $^{2}$\orcidC{}}

\AuthorNames{Andrew Fuchs, Andrea Passarella and Marco Conti}

\AuthorCitation{Fuchs, A.; Passarella, A.; Conti, M.}

\address{%
$^{1}$ \quad Universit\'{a} di Pisa, Department of Computer Science\\
$^{2}$ \quad Institute for Informatics and Telematics (IIT), National Research Council (CNR)}

\corres{Correspondence: andrew.fuchs@phd.unipi.it}

\secondnote{Conceptualization, Andrew Fuchs, Andrea Passarella and Marco Conti; Investigation, Andrew Fuchs; Methodology, Andrew Fuchs, Andrea Passarella and Marco Conti; Software, Andrew Fuchs; Supervision, Andrea Passarella and Marco Conti; Writing – original draft, Andrew Fuchs; Writing – review \& editing, Andrea Passarella and Marco Conti.}

\abstract{Given an increasing prevalence of intelligent systems capable of autonomous actions or augmenting human activities, it is important to consider scenarios in which the human, autonomous system, or both can exhibit failures as a result of one of several contributing factors (e.g., perception). Failures for either humans or autonomous agents can lead to simply a reduced performance level, or a failure can lead to something as severe as injury or death. For our topic, we consider the hybrid human-AI teaming case where a managing agent is tasked with identifying when to perform a delegation assignment and whether the human or autonomous system should gain control. In this context, the manager will estimate its best action based on the likelihood of either (human, autonomous) agent failure as a result of their sensing capabilities and possible deficiencies. We model how the environmental context can contribute to, or exacerbate, the sensing deficiencies. These contexts provide cases where the manager must learn to attribute capabilities to suitability for decision-making. As such, we demonstrate how a Reinforcement Learning manager can correct the context-delegation association and assist the hybrid team of agents in outperforming the behavior of any agent working in isolation.}

\keyword{Simulated Sensing; Entity Detection; Reinforcement Learning; Delegation;} 

\preto{\abstractkeywords}{\nolinenumbers}

\begin{document}

\maketitle

\section{Introduction}\label{sec:introduction}

The prevalence and growing capabilities of autonomous (or human augmenting) systems indicates we can expect an increase in the use of hybrid teams of humans and autonomous systems. While in some cases AI systems can replace human activities, a likely scenario is that of a hybrid system where the human and AI collaborate to solve a certain task. The composition and expectations for these teams will of course depend on the use case and acceptance level for autonomous control. A key aspect of the acceptable level of autonomous control is how reliably the autonomous system can successfully complete the given task. At the very least, to reach acceptance, one would expect a system to meet the performance level of a human. Therefore, it is essential to recognize when an agent can meet this threshold of performance to enable high-performing teams.

In this context, we are not considering a human or AI system's performance at the granular level of individual actions. We consider this a more general measure of desirable behavior as it is less subject to minor details, so our focus can instead prioritize overall success or critical failure. Therefore, our focus is instead on identifying deficiencies which would lead to broader suboptimal performance for humans, AI systems, or both. Recognition of these deficiencies is a key aspect of our context of hybrid teaming. Given the opportunity to delegate tasks to a team member, we should prioritize delegations for members which exhibit fewer performance issues and higher overall performance. Of course, the specific use case will determine the constraints regarding delegations (e.g., frequency, maximum number of assigned tasks, safety requirements, etc.), but the primary component we will consider is recognition of factors contributing to reduced performance. With the ability to recognize these deficiencies leading to reduced performance, we aim to properly utilize individual team members to improve overall team outcomes.

Given a notion of individual performance: success rate, safety level, cost, etc., a key step is to determine how best to assign control in each state or context. This applies more specifically to cases where control is not shared, but instead is assigned to one of the entities in the hybrid team. In such a context, a primary aspect is determining a method of associating the measure of success with delegation decisions. In our context, we will train a delegation manager to find these associations. The delegation manager will be responsible for learning to recognize contexts where the human or AI system indicate potential deficiencies through the outcomes resulting from their delegated control. As the manager learns to recognize these cases, it will associate them with an appropriate delegation decision. In our scenario, we will not assume any preference between delegation of a human or AI system, just a prioritization based on perceived performance with respect to outcomes. The association between contexts and failures will be learned in a Reinforcement Learning framework to enable manager training through reward-based feedback. This ensures we will not need to generate a set of rules governing the manager's behavior; instead, the manager will learn through exploring actions in states.

A further goal of ours is to generate a manager which can make these performance-related determinations and execute only a small number of delegations. In other words, we will attempt to generate a manager which not only identifies a desirable agent, but one capable of making fewer delegation changes. The use of a penalty during training lets us discourage this behavior without requiring any domain-specific upper bound on these switches. Discouraging this behavior is obviously desirable in most cases as frequent delegation changes would make the interactions between the human and autonomous system cumbersome, mentally taxing, and potentially dangerous.

To demonstrate our approach, we will train our manager to make delegation decisions based on representations of agent observations and contexts. The manager will choose, in each state, whether to assign control to either the human or the autonomous agent. To demonstrate the importance of the delegation task, we will provide scenarios in which either one, both, or neither of the agents exhibit signs of sensing deficiencies. We will discuss these deficiencies in later sections, but they are intended to approximate various conditions and errors related to perception and the resulting decisions. These can include issues such as: occlusions, obstructed/broken sensors, adverse lighting/weather conditions, etc. Each of these cases demonstrate a scenario in which human or autonomous sensing can encounter cataclysmic failures. In each context, the manager will focus on the potential existence of sensing deficiencies which could result in a failure case. The notion of failure will depend on the use case, but we will focus on safety as it is of primary relevance. 

\subsection{Driving Safety}\label{sec:driving_safety}

To test and demonstrate the performance of our approach, we consider delegation in the context of driving. We assume vehicle control can be delegated to either the human driver or an autonomous driving system. Given this control paradigm, we will focus on safe navigation in a driving environment. For our scenario, the key aspect of the delegation task is recognition of either agent's sensing deficiencies which impact success likelihoods. In the case of driving, there are numerous aspects which contribute to passenger safety and that of other vehicles, pedestrians, etc. in the environment. These can range from driver attentiveness/awareness, adverse weather and lighting conditions, distractions, and more. Since there are numerous factors which contribute to the safe operation of a motor vehicle, we need to select our primary focus. In the work we present in this paper, we consider safety-related factors which can be attributed to an error in sensing. As an example, consider driving in adverse weather conditions, such as fog. Depending on the severity of the fog, the visible range for object detection can be either marginally or significantly impacted. It seems reasonable to assume that the severity of the adverse conditions and the resulting impact on sensing can be a significant contributing factor in a driving incident.

To justify our consideration of adverse sensing conditions as relevant in the context of vehicle crashes, we investigated the Fatality Analysis Reporting System (FARS) data available via \cite{farsData}. This data is available from the National Center for Statistics and Analysis, an office of the National Highway Traffic Safety Administration. Their resources offer numerous tools and data repositories related to driving incidents, which are available for multiple use cases/approaches. These resources support querying and compiling statistical data from crash reports regarding numerous aspects pertaining to the driver, weather, road, etc. Therefore, we utilized these resources to identify several scenarios/conditions of interest. These illustrate common cases of incident as well as likely contributors to adverse sensing outcomes.

As seen in Table~\ref{tab:fars_data_fog}, the FARS data indicates scenarios where accidents occurred in conditions capable of contributing to impaired sensing. Contexts such as fog, darkness, rain, etc. can all contribute to an increased difficulty of accurate perception. As a clear example, heavy fog would easily impact the distance and clarity possible for perception. Therefore, we will consider several adverse conditions as the basis for our various scenarios.

\begin{table}[H] 
	\caption{Sample crash (filtered by injury-only) data involving adverse conditions. In this case, fog and related weather. \label{tab:fars_data_fog}}
	\newcolumntype{C}{>{\centering\arraybackslash}X}
	\begin{tabularx}{\textwidth}{CCCCCCCCCCCCCC}
		\toprule
		\textbf{Year} & \textbf{Jan} & \textbf{Feb} & \textbf{Mar} & \textbf{Apr} & \textbf{May} & \textbf{Jun} & \textbf{Jul} & \textbf{Aug} & \textbf{Sep} & \textbf{Oct} & \textbf{Nov} & \textbf{Dec} & \textbf{Total}\\
		\midrule
		\textbf{2016} & 542 & 557 & 256 & 197 & 550 & 59 & 526 & 1506 & 445 & 700 & 1185 & 583 & 7106\\
		\textbf{2017} & 1995 & 1415 & 517 & 260 & 239 & 522 & 20 & 478 & 639 & 754 & 447 & 1126 & 8412\\
		\textbf{2018} & 900 & 1011 & 624 & 70 & 280 & 28 & 390 & 645 & 374 & 999 & 1639 & 1825 & 8786\\
		\textbf{2019} & 658 & 1487 & 198 & 134 & 326 & 72 & 60 & 308 & 301 & 566 & 332 & 2700 & 7143\\
		\textbf{2020} & 2050 & 1131 & 623 & 184 & 256 & 111 & 353 & 386 & 812 & 817 & 719 & 800 & 8242\\
		\midrule
		\textbf{Total} & 6146 & 5601 & 2218 & 845 & 1650 & 792 & 1350 & 3324 & 2571 & 3836 & 4322 & 7034 & 39690\\
		\bottomrule
	\end{tabularx}
\end{table}

As indicated by the data, crashes occur in various conditions, including cases of mixed conditions. For example, from Table~\ref{tab:fars_data_fog_and_dark}, we see cases including both darkness and fog. Therefore, we can also consider combinations in our scenarios. Our assumption is that, in many cases, the combined effect of the conditions would be stronger than either would be individually.

\begin{table}[H] 
	\caption{Sample crash (filtered by injury-only) data involving adverse conditions. In this case, a combination of fog (and similar) weather with darkness and possible ambient lighting. \label{tab:fars_data_fog_and_dark}}
	\begin{tabularx}{\textwidth}{CCCCCCCCCCCCCC}
		\toprule
		\textbf{Year} & \textbf{Jan} & \textbf{Feb} & \textbf{Mar} & \textbf{Apr} & \textbf{May} & \textbf{Jun} & \textbf{Jul} & \textbf{Aug} & \textbf{Sep} & \textbf{Oct} & \textbf{Nov} & \textbf{Dec} & \textbf{Total}\\
		\midrule
		\textbf{2016} & 53 & 184 & 235 & 0 & 263 & 22 & 239 & 1031 & 44 & 417 & 636 & 154 & 3278\\
		\textbf{2017} & 1052 & 544 & 117 & 114 & 30 & 121 & 20 & 59 & 76 & 39 & 295 & 610 & 3078\\
		\textbf{2018} & 366 & 567 & 306 & 0 & 257 & 0 & 390 & 284 & 123 & 369 & 897 & 1392 & 4951\\
		\textbf{2019} & 465 & 745 & 15 & 0 & 27 & 0 & 0 & 144 & 188 & 123 & 212 & 1166 & 3086\\
		\textbf{2020} & 800 & 693 & 258 & 127 & 193 & 111 & 112 & 257 & 395 & 458 & 574 & 539 & 4519\\
		\midrule
		\textbf{Total} & 2736 & 2733 & 931 & 241 & 770 & 254 & 762 & 1774 & 827 & 1406 & 2615 & 3861 & 18911\\
		\bottomrule
	\end{tabularx}
\end{table}

In the case of autonomous driving, we were unable to identify a similar resource for extensive historical data. We suspect this is due to several factors such as protection of proprietary systems/data, lack of widespread use of autonomous driving systems, etc. Despite our inability to identify a specific data source, we were able to identify several current research topics relating to sensing in the context of autonomous driving. These topics demonstrate various aspects of the challenges regarding sensing in autonomous driving systems, and they indicate how sensing failures can occur. We discuss these topics further in the related works section (see Section~\ref{sec:related_work}). The key result being we identified additional cases and relevant factors which could contribute to an undesirable outcome in an autonomous driving setting.

\subsection{Key Results}\label{sec:key_results}

We demonstrate the use of augmentation observations in a simulated environment to represent various sensing capabilities and inhibitors. These augmentations enable models of sensor output which link operating contexts and the subsequent impact on sensing. The impacts (e.g., reduced accuracy/range, occluded sensors, etc.) generate contexts with high likelihood of agent failure for activities reliant on perception. The impact on perception is represented for both human and autonomous sensing. In some cases, both agent types would exhibit similar deficiencies for the same augmentation, so we therefore can test various conditions for our hybrid teams. As indicated in our results, the manager recognizes the impact deficiencies have on outcomes to improve delegation choices.

With our proposed manager model, we demonstrate a manager capable of learning delegation policies for various simulated contexts with hybrid teams. In each context, these hybrid teams demonstrate mixed cases where one, both, or neither agent is limited by perception error. In this teaming context, our manager demonstrates an ability to select the appropriate delegations to enable team success. We see our trained manager avoiding collisions whenever there is an agent available with the ability to succeed at the task. Further, we see our manager maintains consistent and strong performance across multiple domains. This strong performance is demonstrated despite our penalty for frequent delegations. These results indicate a manager which can learn key states for delegation changes and avoid burdensome delegation behavior.

We further demonstrate the performance of our manager against a random manager. In this context, we illustrate how a random manager could identify the correct delegations by chance, but with impractical behavior patterns. The random manager would require the ability to make an unacceptable number of delegation decisions at too high a frequency (e.g., one decision every second). Even with this unrealistic allowance, the random manager still fails to outperform our trained manager, which we trained with a penalty for the same frequent delegation changes. We see an increase of up to $127\%$ and $390\%$ in mean episode reward for our environments and the elimination of up to $26$ avoidable collisions when using the learning manager in the same scenario versus the random manager.

\section{Related Work}\label{sec:related_work}

Autonomous control, including driving, poses many challenges and potential failure points. The challenges for an autonomous system vary by context and location. In one aspect, learning models are needed to define/generate the underlying behavior for the autonomous system. The methods for defining and training these models spans multiple topics. This includes concepts regarding whether the system should mimic human cognition, reasoning, performance, etc. Further, these models may need to demonstrate comprehension of the mental state, intentions, and more of humans in the case of interactions or related cases (see \cite{10.1145/3580492} for more details). As demonstrated in \cite{urmson2008autonomous, chen2022milestones, ghorai2022state, badrloo2022image, zhang2023perception}, additional relevant aspects regarding challenges include topics such as perception, navigation, object detection, etc. Each of these factors requires a system to convert sensor data into actions. In driving, these actions are subject to the driving conditions, relative positions of obstacles, and more. These factors determine key decision points and points of failure. As we have noted, the set of possible failures depends on the case, but we can include the case of sensing-related failures. We will include/consider several sensor types and cases commonly observed in autonomous driving, such as RADAR, LiDAR, cameras, etc. \cite{yeong2021sensor}. These methods of autonomous sensing are of immediate interest as they are some of the most commonly used methods, and these sensing systems demonstrate differing capabilities and susceptibility to errors.

In the context of driving, we see existing works more specifically focused on the ability to predict driving outcomes. These include cases of models trained to predict or identify the source of a driving failure given driving-related data such as speed, steering angle, etc. \cite{kuhn2020introspective, hecker2018failure, kuhn2021trajectory, besnier2021triggering, kuhn2020better, zhang2023perception}. These works show an ability to use observations to identify or predict likely failures. This is accomplished via an ability to ingest trajectory, state, or related information, then generate a prediction regarding failure likelihood. Such approaches demonstrate an ability to associate current states with possible outcomes but seem primarily focused on the classification task. It could be worth considering a combination of the predictive nature of these approaches with the safety and delegation approach presented in this paper.

A further relevant topic is the case of route planning in autonomous driving and the potential impact of context on outcome. As seen in \cite{zimmermann2020adaptive, guo2019safe, zhang2023perception, badrloo2022image, rosique2019systematic}, the success rates for perception can be adversely affected by weather, sensor type, and related factors. Something as commonplace as snow/ice could result in a significant impact on camera performance \cite{secci2020failures}. In the simplest sense, consider a case where snow accumulation on the vehicle results in a sensor being blocked. Clearly, the accumulation of snow on, or in front of, a sensor would potentially render an entire portion of the system \emph{sensing region} incapable of providing necessary information. Such a failure in sensing critical to system performance could result in a serious incident.

Specific to the context of autonomous systems, we should also consider the issue of perturbation and related attacks on vision and sensing. As seen in \cite{eykholt2018robust, elsayed2018adversarial, akhtar2021advances, deng2020analysis, cao2019adversarial, zhang2023perception, cao2022emerging}, small (often imperceptible) changes to the appearance of an image result in significant changes in classification. As in \cite{eykholt2018robust}, a small sticker on a stop sign could prevent detection. Such an attack could result in a dangerous situation for a vehicle and its passengers. Similarly, we can consider an autonomous system's ability to properly segment and interpret the sensor output. Segmentation is an important aspect of successful interpretation of visual data, so segmentation failures can lead to catastrophic consequences \cite{zhou2019automated}. A system incorrectly associating sensor input and individual entities could result in a missed detection, which could lead to an incident. For instance, failure in object detection causing a missed traffic signal can lead to dangerous circumstances \cite{rahman2019did}.

For our approach, in addition to cases of sensing failure, we consider the task of delegation \cite{balazadeh2020switch, straitouri2021triage, jacq2022lazy, richards2016delegate, palmer2020assessing, candrian2022rise, 9821063}. We are primarily focused on the manager's ability to take advantage of higher-performing agents in a given context. As in our previous work \cite{9821063}, the goal is a delegation manager which learns a policy enabling intelligent delegation decisions. However, in \cite{9821063}, we focused solely on delegation and not on the manager's ability to associate context-induced failures to desirable delegations. In related delegation use cases, the motivation and method may differ, but a key aspect is identifying a delegation strategy best fitting the given use case. For instance, \cite{straitouri2021triage} demonstrates a method which partially inspired our approach. In their work, they define and train a Reinforcement Learning model which can operate under algorithmic triage. In other words, the model should be able to identify which decisions to delegate to which authority based on a performance/priority measure. Similarly, \cite{balazadeh2020switch} demonstrates a delegation case in which there is a cost associated with several components of the scenario (e.g., cost to make a delegation change, agent usage cost, etc.). In this context, delegation is related to both the level of agent success and the associated costs. In a similar context, the goal can be instead to utilize a simpler and lower cost behavior, when possible, only using a more effective and costly behavior policy when necessary \cite{jacq2022lazy}. This means the delegation would be biased to minimize costs where possible. With regards to when it is best/possible to use the cheaper behavior policy, this can be determined by feedback/reward model capable of signifying low-impact or low-significance states. The model can then learn to utilize the lazy or cheaper model in these states.

Each of the above topics relate to, or inspire, aspects of our method. First, we note the significance of potentially erroneous sensing and its impact on successful/safe driving. In this context, sensing failures will underpin the key aspects of undesirable outcomes. An inability to properly perceive or detect entities in the environment directly impacts an agent's ability to properly act. Regarding the delegation task, our approach will offer the ability to account for agent differences more directly in the decisions and their attribution to outcomes. Our inclusion of agent-specific observations and contextual features allows our managing agent to learn an association between the observations, agent performance, and overall outcomes. Combined, these aspects enable a manager to be better suited to account for variations in agent types to better determine desirable actions. Further, compared to related works, we combine aspects regarding consideration of outcome, sensing deficiencies, and delegation to enable improved team performance.

\section{Delegation Manager}\label{sec:delegation_manager}

\begin{figure}[ht]
    \centering
    \includegraphics[width=0.95\textwidth]{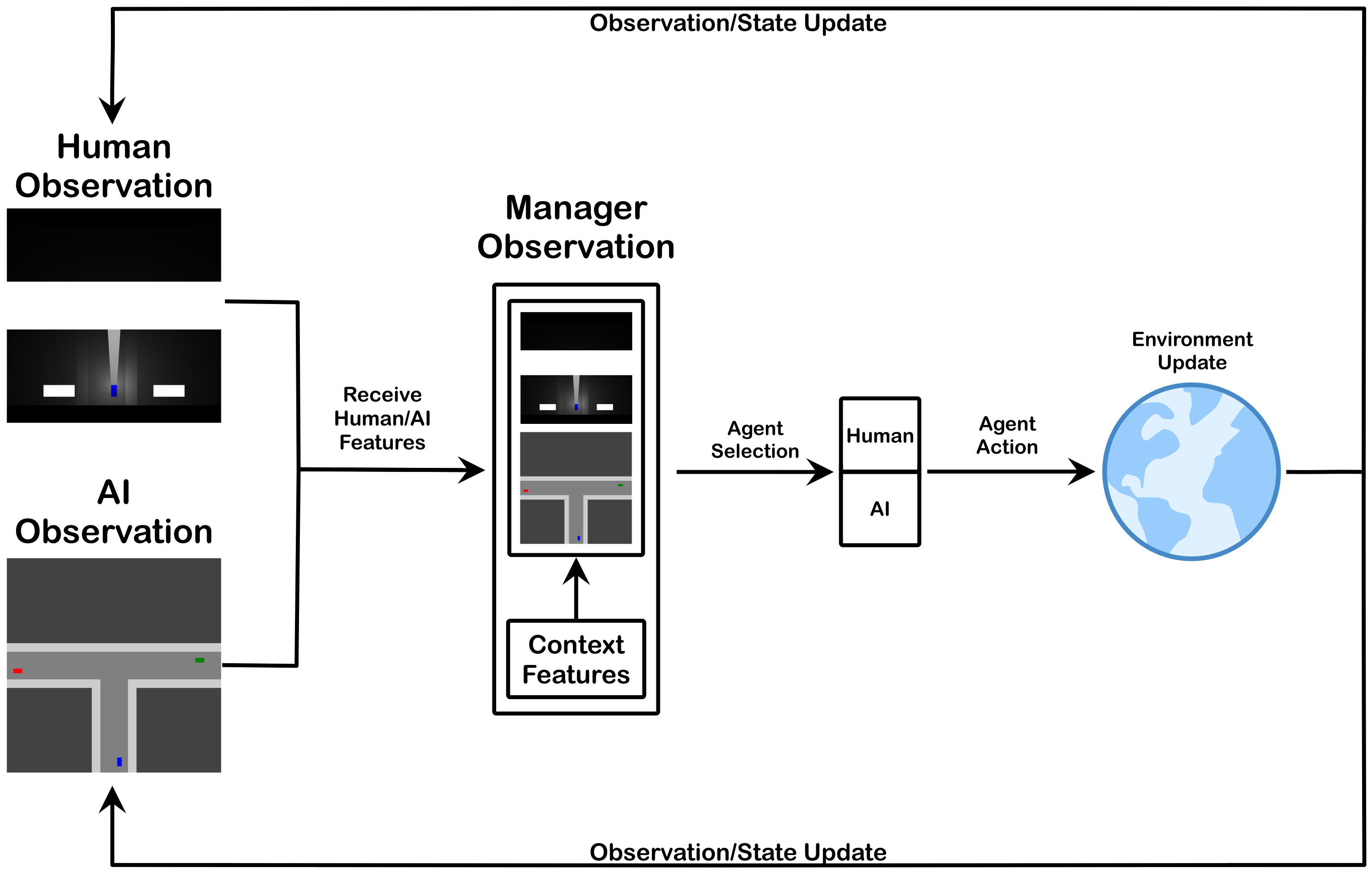}
    \caption{Diagram of manager and agent observation, selection, action, and environment update}
    \label{fig:problem_diagram}
\end{figure}

As noted previously, the delegation task can prioritize various aspects of the problem: agent performance, agent cost, etc. For our purposes, we focus strictly on performance and do not consider cost. Further, we assume no preference between a human or autonomous system having control, just that the delegated decision-maker is the higher-performing of the team. This contrasts with cases of assistive technologies aiding drivers in maintaining safe operation \cite{razak2022modeling}. The delegation decisions are intended for a scenario where the manager is tasked with selecting whether to delegate control to either a human driver or autonomous driving system (see Figure~\ref{fig:problem_diagram}), which we will refer to as the \emph{driver(s)}, \emph{driving agent(s)}, or simply \emph{agent(s)}. The key factor is the manager's ability to recognize which of the two driving agents would be least likely to cause an undesirable outcome (e.g., a collision). As such, we provide a reward signal and training samples to guide the manager's learning. In this driving scenario, the reward function motivates safety from the manager's perspective by penalizing collisions. Consequently, the primary signal indicating desirable delegations is whether the selected driver can safely navigating the given driving conditions. In our scenario, the delegation change should occur when the manager believes there are conditions likely to cause the currently selected driving agent to fail in detecting and responding to another entity in the environment. Our assumption being such a failure would likely lead to a collision when the vehicle paths cross, necessitating a delegation change to avoid the collision.

Regarding the act of delegation, we make some simplifying assumptions. First, we assume both the human and autonomous system are maintaining awareness, to the best of their ability, and are therefore ready to take control when a delegation change is made. Next, we assume the manager can make a delegation decision at any given time step. In other words, there is no explicit restriction on the frequency of the delegations. This is not to say we do not consider how high frequency of delegation changes would be impractical, but instead we simply do not enforce a threshold. In this regard, keeping safety as a key aspect, we prioritize manager behavior which would make the delegation change easier and safer \cite{trosterer2017we}. Therefore, the manager is penalized when it makes frequent/immediate delegation changes. Like how clutter in a head-up display can reduce usability \cite{murali2022intelligent}, we want to ensure our manager maintains usability from the perspective of delegation change frequency. Lastly, we do not assume the manager receives an immediate reward per time step, which would allow them to learn a more direct association between its immediate actions and the feedback received. We instead assume the manager will receive only a single reward at the end of an episode.

We use the single reward to reduce the manager's access to domain-specific information. To define a state-action reward, we would need to use domain knowledge to label the utility of a particular delegation decision in a particular domain/context. By using a single reward, we can define a more general manager training paradigm in which the specific case is irrelevant. The single reward therefore denotes whether an entire sequence of choices results in a desirable outcome, regardless of use case. This will still implicitly enable the manager to generate a relationship between choices and feedback but will significantly reduce the amount of information we assume the manager has access to at training time. In other words, a delegation decision is not immediately rated for the manager but is instead learned through observing outcomes of trajectories (with potentially many delegation decisions). Again, this allows us to significantly reduce the amount of domain knowledge we assume the manager can access. Additionally, the use of a single reward value for all actions in a trajectory prevents training a manager which exploits particular utilities for specific states. Therefore, like the chess example in Section 3.2 of \cite{sutton2018reinforcement}, the manager must learn to improve trajectories/outcomes rather than trying to reach a particular state or achieve a subgoal. Note that, in principle, this approach requires a combinatorial number of trajectories to train the manager in each specific action. However, the approach also allows the manager to learn combinations of actions and their dependencies more generally.

The aforementioned characteristics denote the key aspects of the manager's goal and responsibility. As seen in Figure~\ref{fig:problem_diagram}, the manager provides oversight to make performance-improving delegation decisions. This starts with the manager receiving state information from the human and autonomous system and additional contextual details (e.g., weather, lighting, distance/direction to goal, etc.). Based on these states and contexts, the manager decides on the current delegation. In the first state of the episode, this is a selection of the agent who will perform the first action. In all subsequent states, the manager makes a similar determination, but is instead deciding whether to keep the current agent or change the delegated agent. In either case, the manager is using an association between state/context information and corresponding driver performance to learn the most suitable delegations and when to make them. Each delegation decision gives authority to a particular driver, and the driver's subsequent action generates an updated state for the hybrid team. This update is reflected in an environment update based on the vehicle moving along its path. The environment updates are then observed, with the imposed agent-specific perception constraints, by each agent and the cycle repeats.

\subsection{Manager Model}
\begin{figure}[ht]
	\centering
	\includegraphics[width=0.7\textwidth]{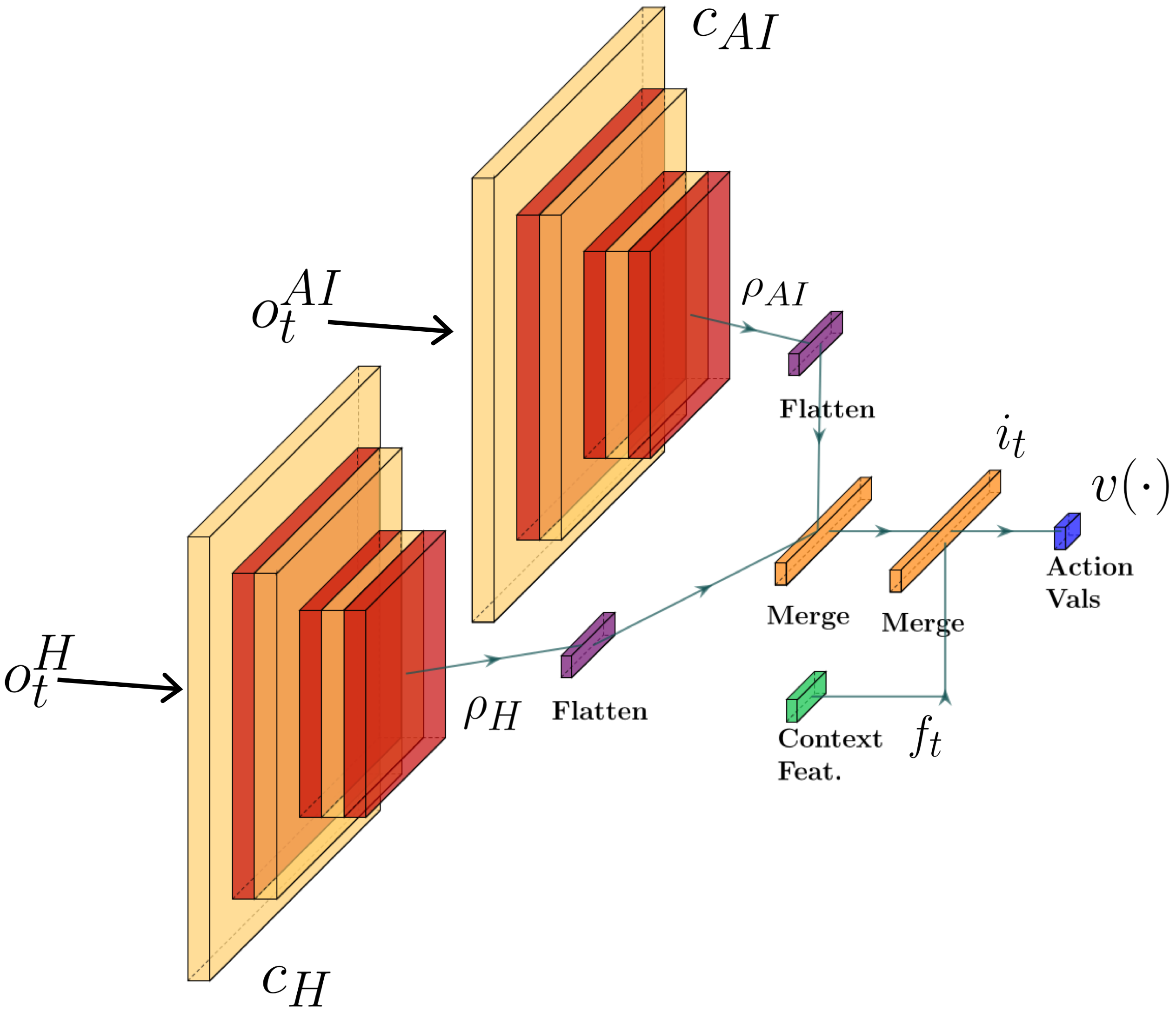}
	\caption{Manager Deep Reinforcement Learning Architecture Diagram}
	\label{fig:manager_diagram}
\end{figure}

To define the manager, we utilize an architecture capable of integrating both visual and contextual input to learn a behavior policy. The manager is expected to observe representations of state for the human and autonomous system, along with their corresponding contextual information. As such, we utilize two Convolutional Neural Network (CNN) heads to ingest the state representations (human and autonomous driver, respectively). The output values of the two CNN heads are then concatenated along with a vector representing additional contextual information (e.g., positional data). Finally, these values are passed to the remaining hidden layers forming our Deep Q-Network as seen in Figure~\ref{fig:manager_diagram}. This architecture was chosen as it enables several aspects of the learning process. First, the use of CNN heads enables direct manager observation of (visual/sensor) perception estimates. Further, the use of two heads enables a separate association for human perception and AI perception. Lastly, the use of a Deep Q-Network architecture enables manager training through reward-based feedback during exploration of delegation decisions in the given state spaces (i.e., driving context). This allows the manager to learn an association between states/contexts and delegation actions.

Following the diagram of observation, delegation, and action flow seen in Figure~\ref{fig:problem_diagram} and the network diagram seen in Figure~\ref{fig:manager_diagram}, the process of converting observations to manager actions are as follows. First, at time $t$, the human and AI agent provide their observations $o^H_t$ and $o^{AI}_t$, respectively. In our scenario, these observations are RGB images with height and width assigned per agent based on sensing capabilities. The manager has CNN heads $c_H$ and $c_{AI}$ which receive the corresponding human and AI observations as input. Each head $c_x, x\in \{H, AI\}$ is comprised of several convolutional layers. The output of each layer undergoes Batch Normalization \cite{bjorck2018understanding} and is passed to the layer's activation function. In our model we use the rectified linear unit \cite{nair2010rectified}
\begin{equation}\label{eqn:relu}
    g(z) = \max(0, z)
\end{equation}
for activation. The output of the final layer of a CNN head, $\rho_x = c_x(o^x_t)$, is then flattened to create a compatible input vector for our remaining hidden layers.

Prior to the hidden layers, the $\rho_x$ are concatenated to generate a single vector $\rho_t = [\rho^T_H, \rho^T_{AI}]^T$ representing the processed visual model of the current state for the team. As additional input, the contextual features (e.g., weather, light, goal distance, etc.) are converted to vector form $f_t\in\mathbb{R}^k$, where $k$ depends on the context features supported. For our model, the elements of vector $f_t$ are the (normalized) scalar representation of the contextual features. The final combined vector $i_t = [\rho^T_t, f^T_t]^T$ serves as input to the remaining hidden layers of our Deep Q-Network. As output, the Deep Q-Network provides a value estimation $v(\cdot) \in\mathbb{R}^{|A|}$, where $|A|$ denotes the size of the manager's action space. Finally, our manager chooses the delegation $\delta_t$ for the current state by selecting the $\argmax$ action:
\begin{equation}\label{eqn:manager_value}
    \delta_t = \argmax_{a\in A} v(a)
\end{equation}

To motivate desirable manager delegation choices, we include several factors in the manager reward function. First, we want delegations to lead to efficient paths from the start to goal positions. Next, we want the manager to learn a strong motivation to avoid vehicle collisions, so we include a large penalty for such cases. Further, we want the manager to avoid making many immediate delegation switches. This is intended to improve the ease of use as it should prevent the manager from frequently, and within short time windows, making multiple delegation changes. Such delegation changes would be challenging for a human user as the manager would appear erratic and require increased attention. Additionally, a manager making frequent changes would increase the likelihood of encountering erroneous behavior while the control is transferred to/from the human driver. Lastly, we want to provide the manager with a positive reward for a successful navigation to a goal state as this is the underlying behavior we are trying to accomplish with all drivers. Therefore, we define the manager reward as
\begin{equation}\label{eqn:manager_reward}
	R=\begin{cases}
		100 - s_e - d_e \quad &\, \text{goal is reached} \\
		-100 - s_e - d_e \quad &\, \text{otherwise} \\
	\end{cases}
\end{equation}
where $s_e$ denotes the number of steps to reach the goal and $d_e$ the number immediate delegation reversions in an episode. In other words, $d_e$ signifies the number of times the manager made two delegation changes in a row. We use $d_e$ as a penalty to discourage these immediate delegation changes as they would be inefficient and unwieldy in many scenarios. We include the additional penalty $s_e$ as further motivation for the manager. In this case, it is a motivation to identify the more efficient of the two driver behaviors with respect to time, so efficient driving is prioritized (within the confines of safety considerations).

\section{Behavior Models Related to the Driving Case}\label{sec:driving_behavior}

To provide samples for the training manager, we control several aspects of the driving behavior. As one important aspect of our goal was to indicate the impact of sensing on driving, we prioritized the corresponding aspects behavior. To accomplish this, the vehicles are provided predetermined paths through the environment, but the behavior models can make collision-avoidance corrections by decelerating to allow the other vehicle to bypass or lead the primary vehicle. This then places the key aspect of collision avoidance on sensing and detection. On the other hand, other aspects of successful navigation (e.g., lane departure) are omitted.

The paths vehicles follow are generated as a shortest path along a directed graph from the vehicle's starting position to its goal. We generate the graphs based on the road configuration in a two-lane road scenario. The use of bidirectional traffic enables environments with more realistic cases where collisions are possible. These include cases where vehicles are both attempting to make turns where their trajectories would cross. Additionally, this also allows cases where two vehicles would occupy the same road, but in opposite directions. Such a case would result in a detection of another vehicle, but not require any response as the paths would not cross given both vehicles drive straight. To perform the driving task, vehicles follow their given paths without deviation via sudden/random changes in steering.

To prioritize the impact of sensing, the driving behavior models are only allowed to make decisions regarding vehicle acceleration. In the case where no obstacles or collisions are detected, the vehicles will accelerate to the prescribed maximum velocity for their current state. These maximum velocities are intended to provide a reasonably accurate representation of real driving speeds when in a $50 km/h$ area (e.g., reduced speeds when turning). The following parameters were used to constrain the vehicle dynamics:

\begin{table}[H] 
    \caption{Parameters used to constrain driving behaviors. \label{tab:driving_parameters_and_thresholds}}
    \begin{tabularx}{\textwidth}{CC}
        \toprule
        \textbf{Parameter}	& \textbf{Value}\\
        \midrule
        Minimum Safe Distance & $10$ meters\\
        Minimum Deceleration Magnitude & $0.2 m/s^2$\\
        Maximum Acceleration Magnitude & $1.2 m/s^2$\\
        Goal Reached Threshold & $1$ meter\\
        Maximum Driving Speed & $13.5 m/s$\\
        Maximum Left Turn Speed & $5.5 m/s$\\
        Maximum Right Turn Speed & $4.2 m/s$\\
        \bottomrule
    \end{tabularx}
\end{table}
The intention for such a restriction on the behavior is that the ability to detect another vehicle, and decelerate accordingly, will rely on the sensing capabilities of the human or autonomous system. As such, erroneous behavior leading to a collision should result from a failure to detect other vehicles rather than an ability to understanding basic driving behavior.

To distinguish behavior between the human driver and autonomous system, we will alter their sensing capabilities individually. In other words, the two driving behavior policies will not differ in their ability to navigate the road and make acceleration decisions; instead, they will differ in sensing/perception. This is a reasonable assumption as we would expect a deployed autonomous system to perform at levels meeting, or exceeding, that of its human counterpart. Therefore, the success of a driver and its corresponding behavior is distinguishable by its corresponding sensing capabilities. For instance, we could define a case where the observation range for the human is half that of an autonomous system, but all other sensing is the same. This basic example would generate two behavior models as their decisions would be based on distinct views of the world. A more extreme case would be a human showing restricted visual perception due to environmental factors (e.g., fog) while the autonomous driving system can perceive the environment as if the fog were not present. This sensing disparity would create a major separation in entity detection ability, which would again lead to a notable change in agent behavior.

Detection of other vehicles is of course a key aspect of collision avoidance. To successfully navigate an environment with other vehicles, the driver must detect and avoid potential collisions. Our collision avoidance can be separated into two primary cases. First, the case where one vehicle is following a lead vehicle while traveling faster than the lead. In this case, the trailing vehicle should be expected to slow down to avoid a rear collision with the lead vehicle. For the second case, we consider the scenario where one or both vehicles turning would result in intersecting paths. In this case, one of the vehicles should relinquish the right-of-way and allow the other vehicle to proceed first. In both cases, we ensure that there is always one vehicle allowed to proceed as intended, which prevents cases where all vehicles are stuck while each try to avoid the others. As is clear in both cases, the ability to respond to such scenarios will rely on the ability to detect them. Given the method of relinquishing right-of-way, our collision avoidance then comes down to a matter of sufficient detection.

In our scenario, we further divide the collision cases by entity type. For our experiments, we are utilizing a single managed vehicle, so all remaining vehicles in the environment are considered background vehicles. These background vehicles are not being tested for their ability to avoid collisions, so they are not given any significant restrictions on sensing which would lead to collisions. However, these vehicles are still responsible for avoiding collisions with each other. Therefore, they do not consider the existence of the managed vehicle, which places the onus for collision avoidance entirely on the managed vehicle when it is in a potential collision scenario. In effect, this allows us to maintain a collection of potential vehicles for interaction while also forcing the managed vehicle to identify a suitable course of action. Further, this approach prevents background vehicles from generating false collision cases from the manager's perspective.

\subsection{Sensing Contexts}\label{sec:sensing_contexts}

As the basis for our sensing representation, we assume each driver has parameters defining the constraints of their sensing capabilities. The first is the observation distance in the four basic directions (forward, backward, left, and right). These define the boundaries of the \emph{sensing region} for each driver. This sensing region is what can be modified further by the contextual details considered above (e.g., fogginess). The modification of the observations allows us to generate various contexts with diverse levels of severity. Continuing with the fogginess example, this would refer to how dense the fog is and how quickly the density increases as a function of distance from the driver's position. In the following sections, we will detail further the contexts we consider and their corresponding significance.

\begin{figure}[ht]
    \centering
    \begin{subfigure}[t]{0.23\textwidth}
        \centering
        \includegraphics[width=\textwidth]{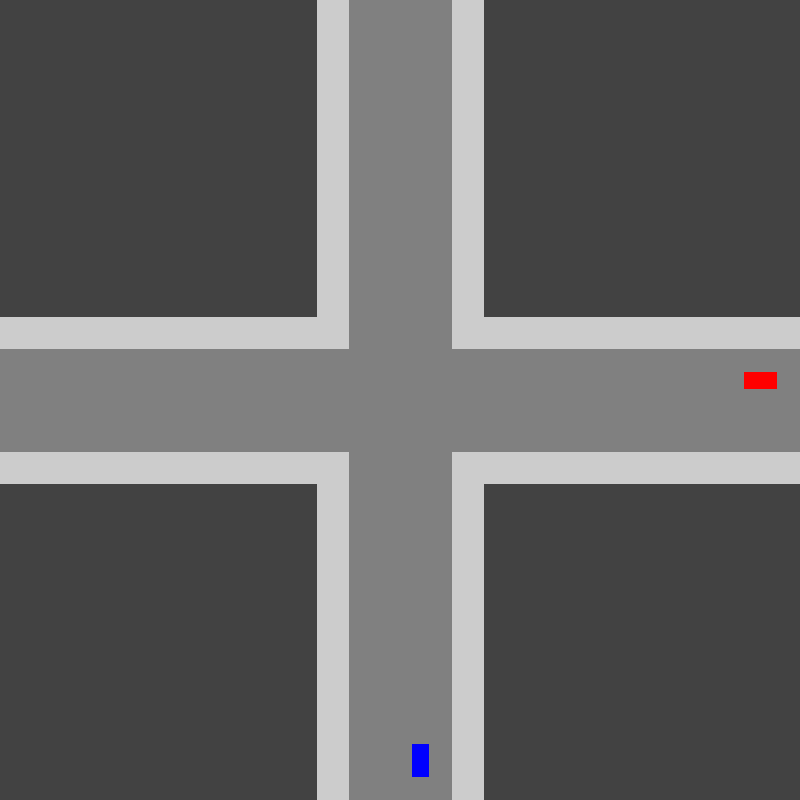}
        \caption{Base case}
        \label{fig:no_context}
    \end{subfigure}
    \hfill
    \begin{subfigure}[t]{0.23\textwidth}
        \centering
        \includegraphics[width=\textwidth]{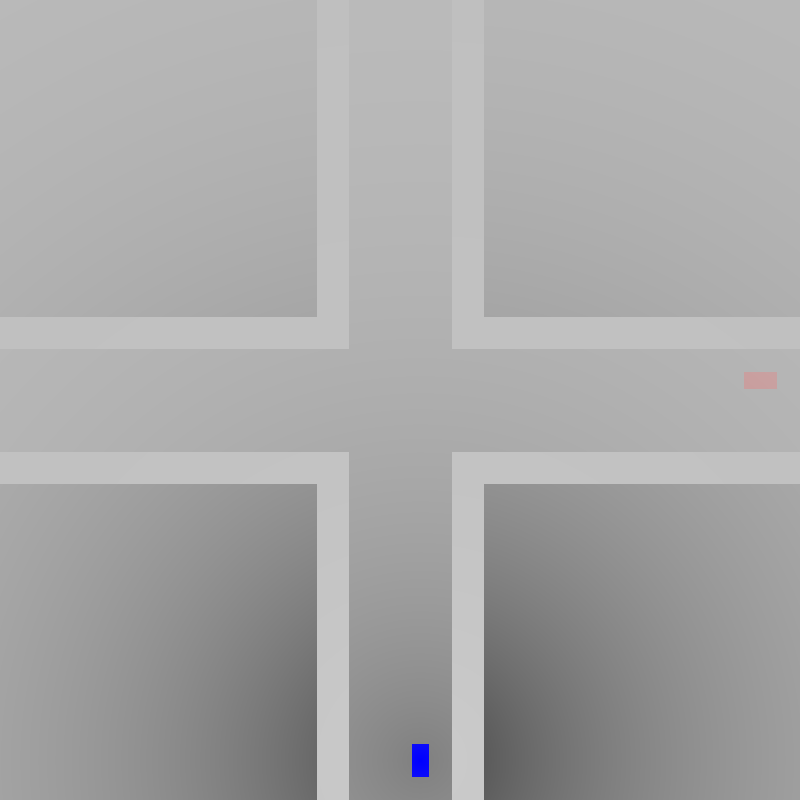}
        \caption{Foggy weather}
        \label{fig:fog_context}
    \end{subfigure}
    \hfill
    \begin{subfigure}[t]{0.23\textwidth}
        \centering
        \includegraphics[width=\textwidth]{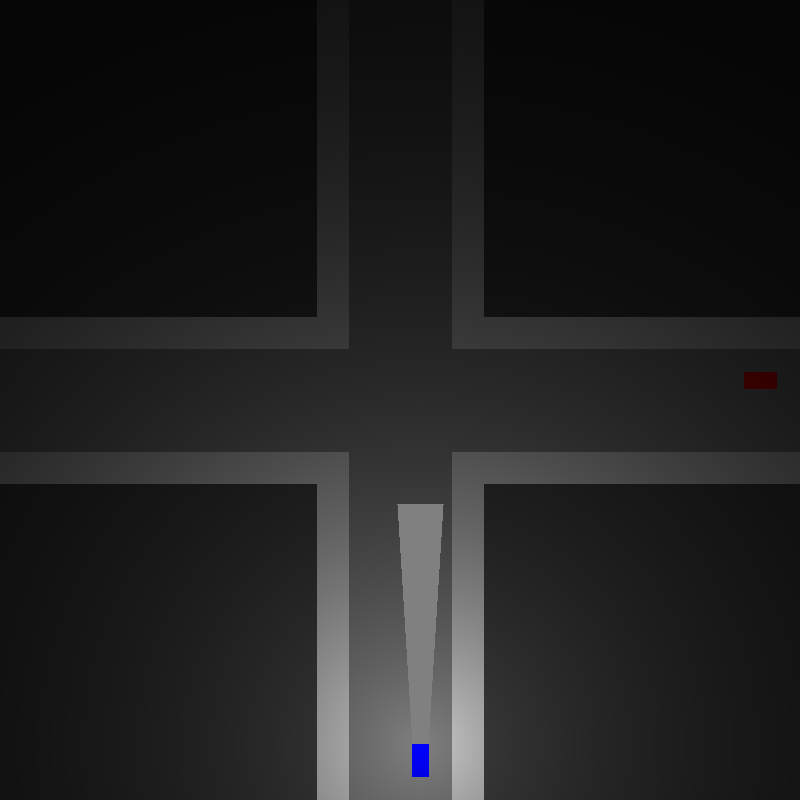}
        \caption{Night/Low light}
        \label{fig:night_context}
    \end{subfigure}
    \hfill
    \begin{subfigure}[t]{0.23\textwidth}
        \centering
        \includegraphics[width=\textwidth]{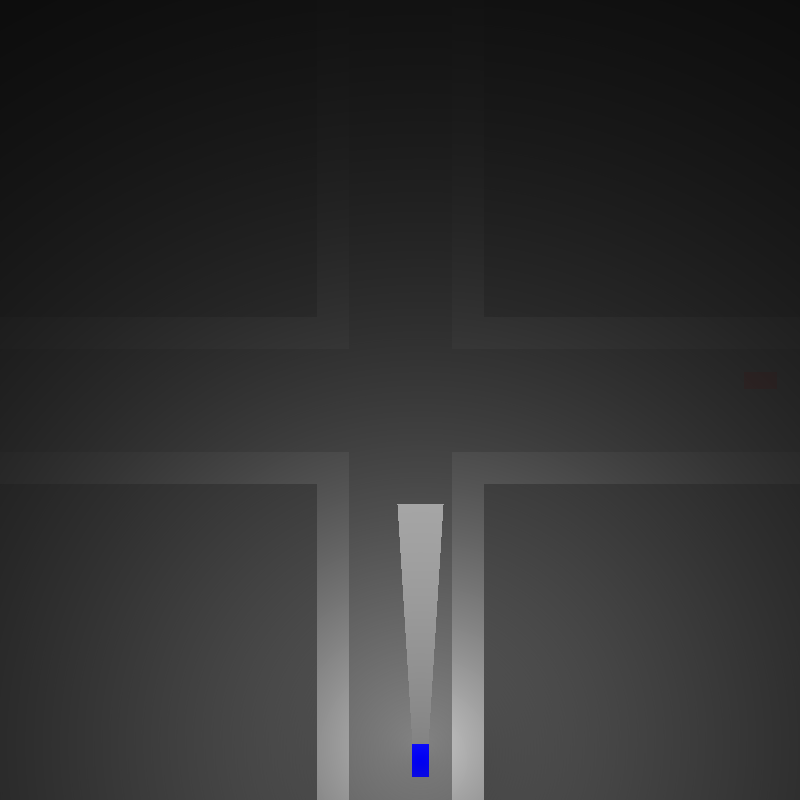}
        \caption{Night with fogginess}
        \label{fig:fog_night_context}
    \end{subfigure}
    \caption{Adverse driving condition contexts.}
    \label{fig:weather_light_context}
\end{figure}

\subsubsection{Obstructed and Failed Sensing}\label{sec:failed_sensing}

\begin{figure}[ht]
    \centering
    \begin{subfigure}[t]{0.23\textwidth}
        \centering
        \includegraphics[width=\textwidth]{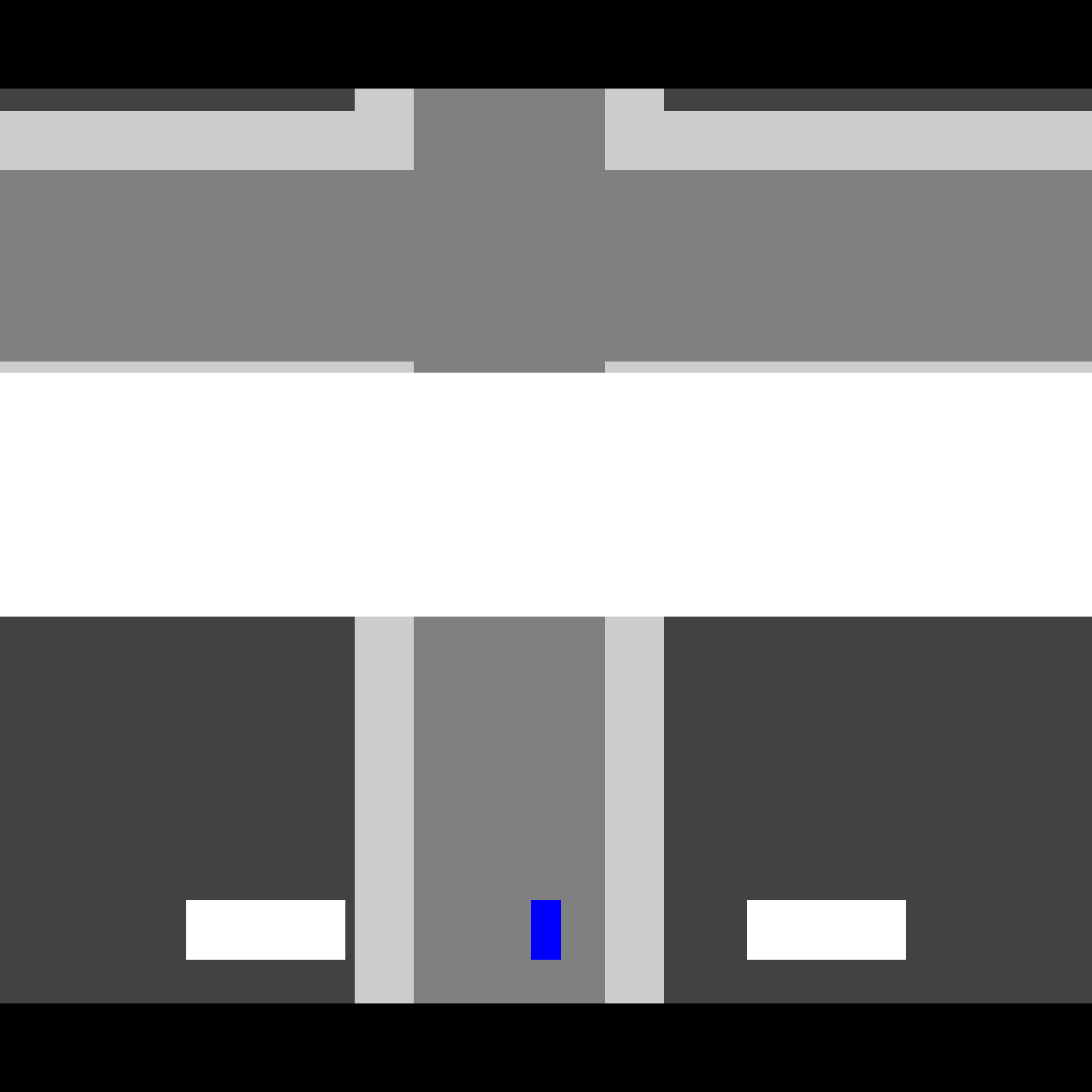}
        \caption{Base sensing case}
        \label{fig:sensing_no_context}
    \end{subfigure}
    \hfill
    \begin{subfigure}[t]{0.23\textwidth}
        \centering
        \includegraphics[width=\textwidth]{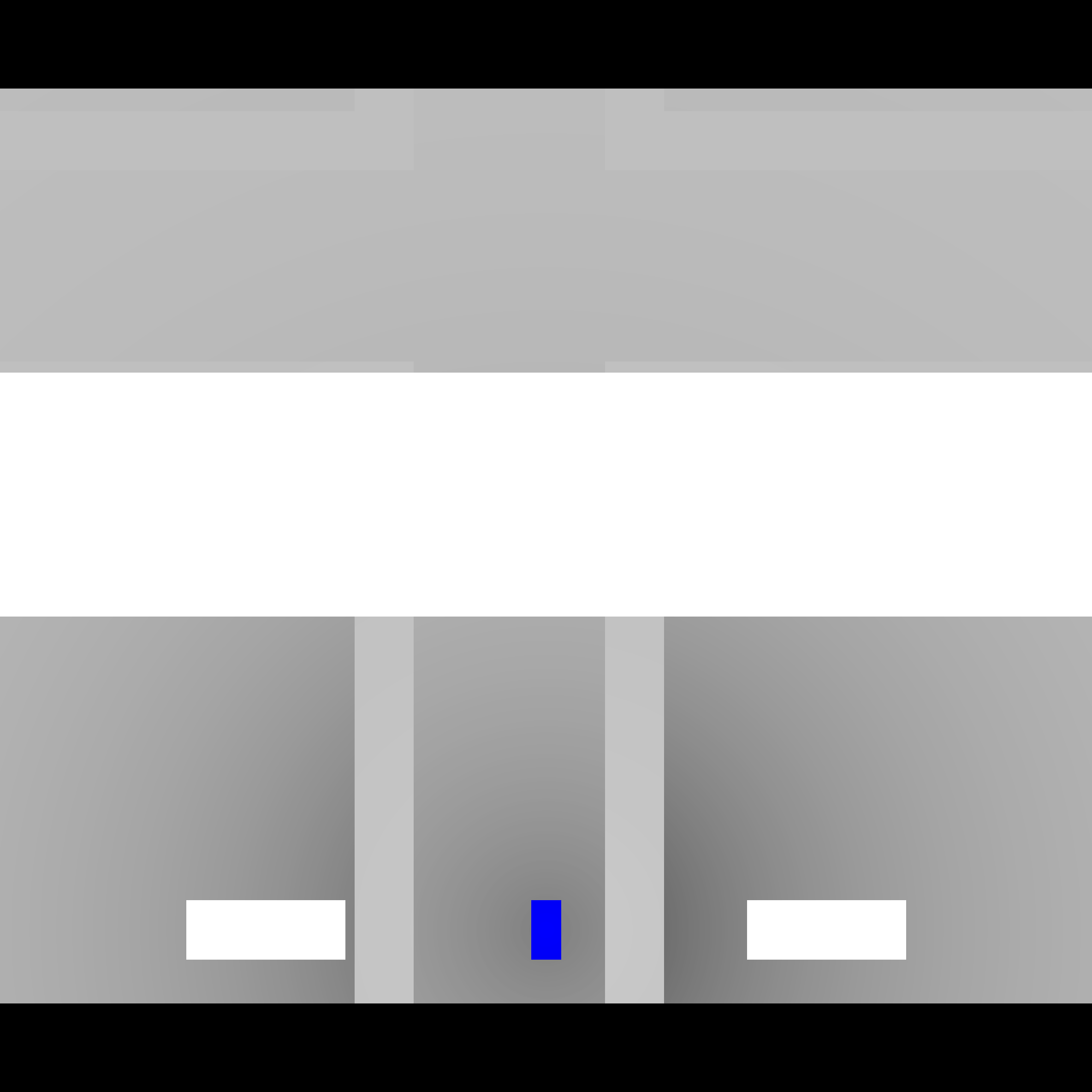}
        \caption{Sensing in foggy weather}
        \label{fig:sensing_fog_context}
    \end{subfigure}
    \hfill
    \begin{subfigure}[t]{0.23\textwidth}
        \centering
        \includegraphics[width=\textwidth]{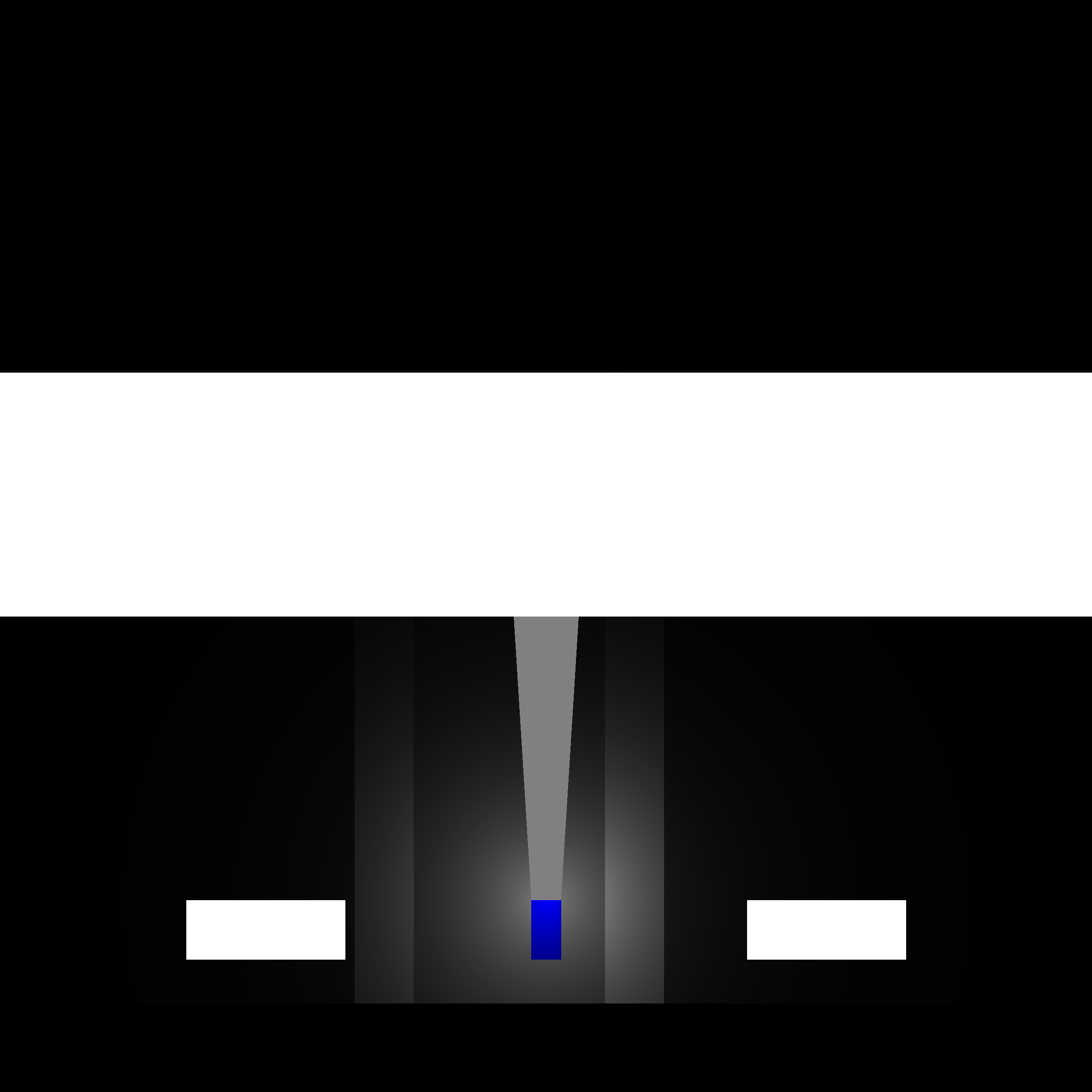}
        \caption{sensing night/low light}
        \label{fig:sensing_night_context}
    \end{subfigure}
    \hfill
    \begin{subfigure}[t]{0.23\textwidth}
        \centering
        \includegraphics[width=\textwidth]{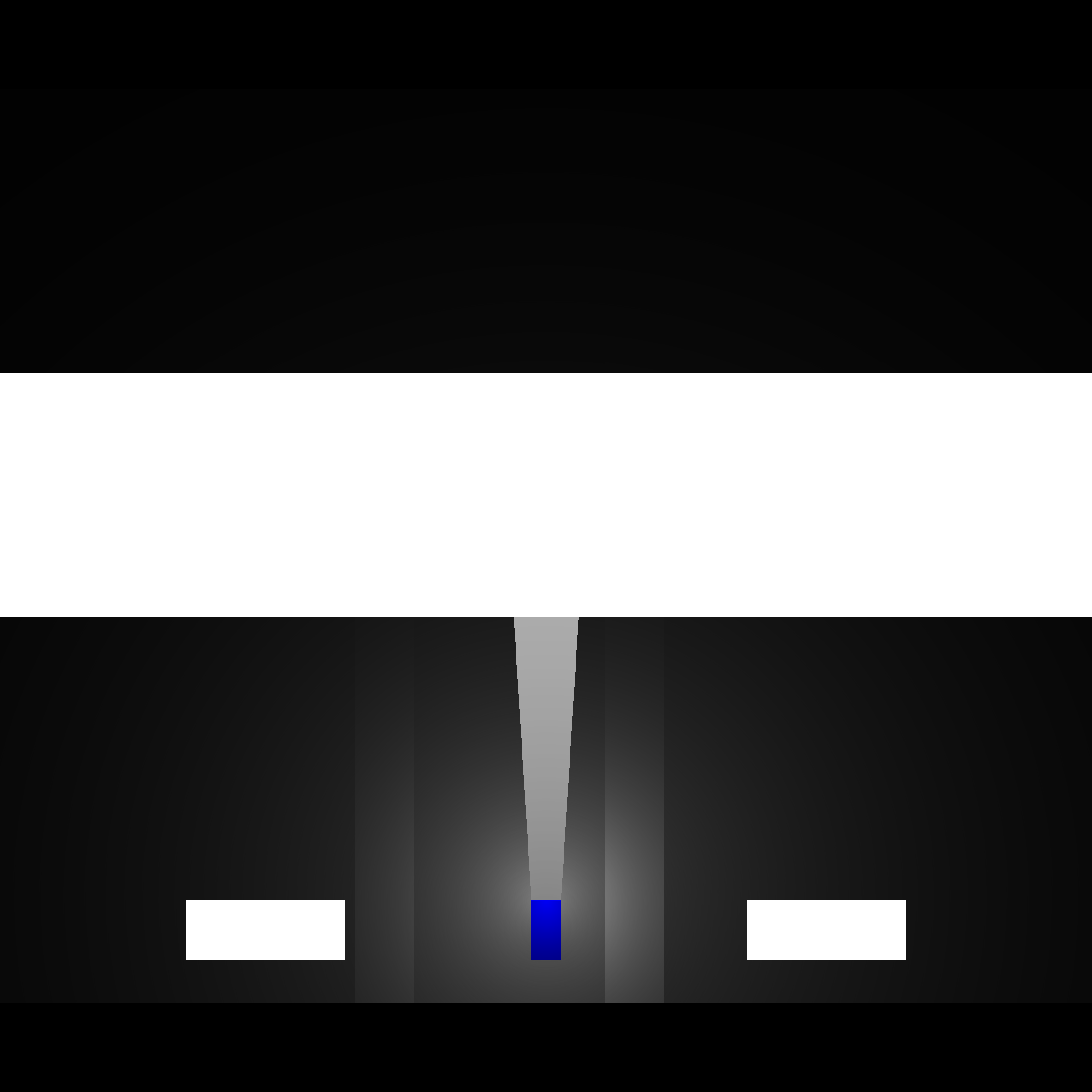}
        \caption{Sensing night with fogginess}
        \label{fig:sensing_fog_night_context}
    \end{subfigure}
    \caption{Sensing with masks for adverse driving condition contexts.}
    \label{fig:sensing_weather_light_context}
\end{figure}

In obstructed sensing case, we consider scenarios where the human or autonomous system have areas of the sensing region blocked from their view. This could be something as simple as a blind spot for human drivers or as significant as a sensor failure in an autonomous system. For either driver type, it is apparent there will be regions of their environment which go unobserved. In our context, we will refer to the models of these obstructed areas as \emph{masks} (see Figure~\ref{fig:sensing_no_context}). The masks will represent an inability to observe the region covered by the mask, which can subsequently lead to situations with failed detection resulting in a collision. This is most simply demonstrated in cases where the collision vehicle is completely obscured by the mask. In such a case, the driver would never observe the vehicle, and a collision for that driver would be unavoidable. In a less extreme case, the mask(s) may obstruct only a portion of the region the collision vehicle will occupy. In this case, the driver has a chance to observe the vehicle. In this case, the driver has diminished perception of the other entity for a certain time window, then the entity would leave the masked region and become visible to the driver.

In our simulation, the severity of a mask's impact in a sensing region is based on how extensively the mask reduces the visible area of the entity. As we are operating in a two-dimensional scenario, the reduction in visibility is represented by a mask covering an entity from an aerial perspective. If the area covered by the mask is $50\%$ of the agent's visible area, then the detection likelihood would be degraded the same amount. If there is no region of the entity covered by the mask, then the detection in the current state will continue as if no masks exist for that agent's detection likelihood. This allows us to consistently measure and enforce the impact masks will have on detection. The area of a given mask is given as a configuration parameter.

For each driving agent, their observation instance is provided with parameters denoting which of the four sensing directions will have a mask, where the masks will be positioned in the sensing region, and how large they are relative to the sensing region. As seen in Figure~\ref{fig:sensing_no_context}, the agent has three masks visible in the observation. The large white rectangle in front of the vehicle represents a mask covering the full width of the front sensing region, but only a portion of the depth. Additionally, two additional masks are given for the left and right sensing regions. With respect to the relative position, we support either: 1) adjacent to the vehicle; 2) centered in the sensing region; or 3) adjacent to the sensing boundary. These parameters allow us to define one or more masks for a driving agent, each with varying levels of severity.

\subsubsection{Weather (Fogginess)}\label{sec:weather}

Another scenario we support in the observation simulation is foggy weather. In this case, we generate a representation of fog via a convex combination of two images. The first image is the original scene without weather context. The second image has equal dimensions to the first, but with all pixels set to the color gray. The exact color of gray can be set via an additional parameter, but we use the RGB values $(191, 191, 191)$ for a soft gray. The convex combination is done per pixel based on an exponential decay function.
\begin{equation}\label{eqn:exp_decay}
    f(x) = e^{-x/\alpha}
\end{equation}
where $\alpha$ determines how severely the function decays, and where $x$ is the distance in pixel space between the point of observation and the coordinate observed. The value $\alpha$ is defined by two parameters $\delta$ and $\gamma$. These denote the distance to the severity point and the decay value at the severity point, respectively. In other words, $\delta$ and $\gamma$ determine how far from the observation point the fog level will reach a given severity level. These values then define $\alpha$ as expected:
\begin{equation}\label{eqn:alpha}
    \alpha = \frac{-\delta}{\ln{\gamma}}
\end{equation}
Given Equation~\ref{eqn:exp_decay} and the parameter $\alpha$, from observation center $c = (x, y)$ we generate foggy image $F$ by performing the convex combination of original image $I$ and gray image $G$:
\begin{equation}\label{eqn:fog_image}
    F_{ijk} = f\left(d(c, (i, j))\right)I_{ijk} + \left[1 - f\left(d(c, (i, j))\right)\right]G_{ijk}
\end{equation}
where $d(\cdot, \cdot)$ is the Euclidean distance. Using this method, we generate a simulation of fog in the observations (see Figure~\ref{fig:fog_context}) from an agent's perspective.

Like the masked sensing case, we use the context to degrade the driver's ability to detect an entity. In the fogginess case, we degrade the detection likelihood based on the severity of the fog at the location of the entity. The mean fog weight value for the region occupied by the entity is the coefficient used to scale the detection likelihood. As noted above, the detection likelihood is based on the ratio of observable area of an entity versus their total area. This ratio is therefore reduced further by the fog-based weight to generate a weather-based impact on detection for the driver.

\subsubsection{Light (Nighttime and Daytime Driving)}\label{sec:light}

As in the fog scenario, we can expect driving scenarios to include cases where the environment conditions vary by time of day and the level of visibility determined by the light levels. In this context, there are some overlaps with the fog case. We would expect a similar sort of decay in the sensing ability as the darkness level degrades the ability to perceive the other entity as distance or severity increases. One clear difference is of course the inclusion of headlights. In the case of night driving, unlike the fog case, the area impacted by the headlights is of course a region with a much easier task of detection. Given these key similarities, we use methods seen in the fog case when modifying detection likelihood.

As seen in the fog case, we need to degrade the base observation output of the simulator to generate a scene mimicking a nighttime driving scenario. Fortunately, this can be accomplished similarly. In our method, we again assume visibility decay but use black instead of gray to increase darkness as distance increases from the observation point. This allows for another predictable level of degradation as distance increases while also giving a reasonable analog to more realistic lighting conditions. We do not claim this perfectly represents real-world night conditions, but it should serve as a reasonable demonstration.

In addition to degrading the color of the pixels as a function of distance, we need a method for generating the headlight region. In our approach, we define the headlight region via several parameters: headlight region depth, car width, and expansion angle. The region depth of course refers to how far from the front of the vehicle the headlight region should expand, and the car width determines the width of the region directly adjacent to the vehicle front. The last parameter, expansion angle, will determine how much the region expands as you move away from the vehicle. These parameters generate a trapezoidal region which simulates light altering the observation. For our purposes, we assume that any pixels under this region will be treated the same as daytime levels. Therefore, any pixels outside the headlight region will undergo the light-based decay and those within the headlight region are unaltered (see Figure~\ref{fig:night_context}).

\subsubsection{Color-based Detection Failure}\label{sec:color}

For a case more specifically focused on errors in autonomous driving, we included an approach representing detection errors specific to Machine Learning and Artificial Intelligence topics. More specifically, we consider scenarios where the autonomous system could fail to properly detect or identify an entity due to an issue in the detection or recognition algorithm(s). These errors are related to topics such as image segmentation and perturbation attacks. It has been demonstrated that vision-based AI systems are susceptible to errors which can have a significant impact on performance \cite{secci2020failures, eykholt2018robust, zhou2019automated, cao2019adversarial} in autonomous driving, so our goal was to demonstrate these in our simulation.

We use a single method as analog for such detection and recognition errors described above. Namely, we support the inclusion of a color parameter which signifies the color a driver would be unable to properly interpret. In one sense, this could be viewed as the driver failing to detect objects of a certain color. Such a case would mean the detection system is either completely blind to entities with this color, or more realistically, the system is unable to properly recognize them as distinct entities. Such failures could result in entities being grouped with others when segmenting the observation. In this case, we are assuming the segmentation failure would result in the agent either ignoring or misinterpreting the danger of another entity. In the other sense, we can view this error type as the improper recognition of entities having a particular property. This is more closely related to topics such as perturbation attacks where the classification of an entity is altered by a slight change in its appearance. These slight changes can significantly impact the outcome. For instance, a small sticker on a stop sign could prevent a proper recognition of the sign.

Such an outcome from either interpretation could obviously be catastrophic, so we use this as the inspiration for our color-based context representation. In our case, we allow setting any colors which should result in a detection failure. Again, in its most basic sense, this would mean the agent cannot properly recognize an entity's existence, resulting in a failure to respond. To support this approach, we include methods for measuring color similarity and determining the corresponding impact on detection. To measure color similarity, we utilize a weighted Euclidean distance as seen in \cite{www_riemersma_2019}. For colors $C_1 = (R_1, G_1, B_1)$ and $C_2 = (R_2, G_2, B_2)$, the color distance is defined as:
\begin{equation}\label{eqn:color_distance}
    d(C_1, C_2) = \sqrt{\left(2 + \frac{\bar{r}}{256}\right)(R_1 - R_2)^2 + 4(G_1 - G_2)^2 + \left(2 + \frac{255 - \bar{r}}{256}\right)(B_1 - B_2)^2}
\end{equation}
where $\bar{r} = \frac{1}{2}(R_1 + R_2)$. This metric attempts to include a correction for brightness perception. This version approximates related methods used in color models. Given a measure of color similarity, we can therefore create a weight for how likely a driving agent is to recognize an entity based on its corresponding color. The severity is the ratio of the color distance and the maximum color distance. In other words, a greater difference in color results in a smaller (or no) impact on detection failure.

\begin{figure}[ht]
    \centering
    \begin{subfigure}[t]{0.23\textwidth}
        \centering
        \includegraphics[width=\textwidth]{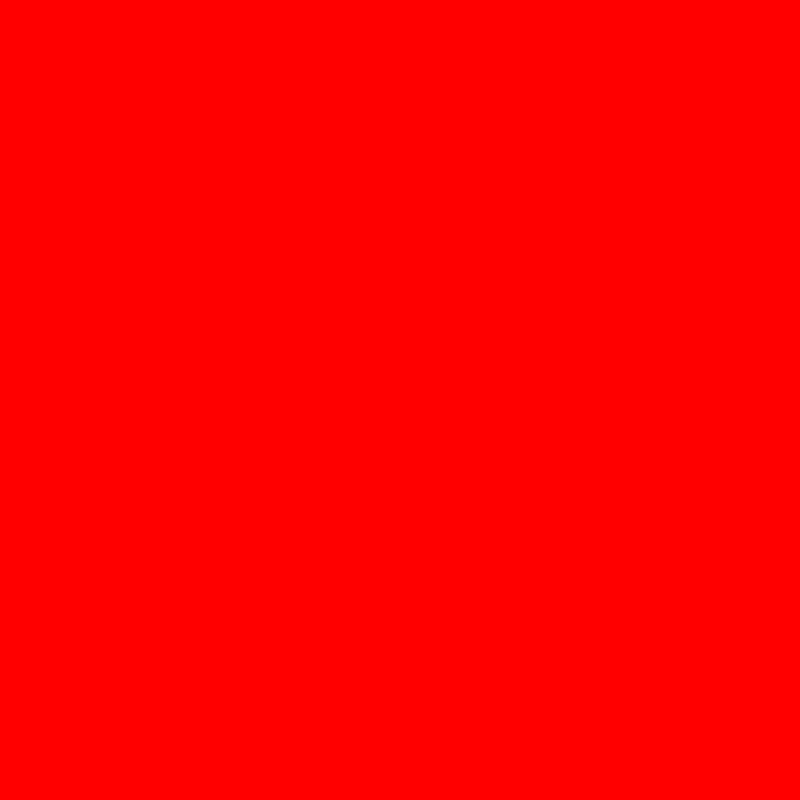}
        \caption{Default}
        \label{fig:color_no_context}
    \end{subfigure}
    \hfill
    \begin{subfigure}[t]{0.23\textwidth}
        \centering
        \includegraphics[width=\textwidth]{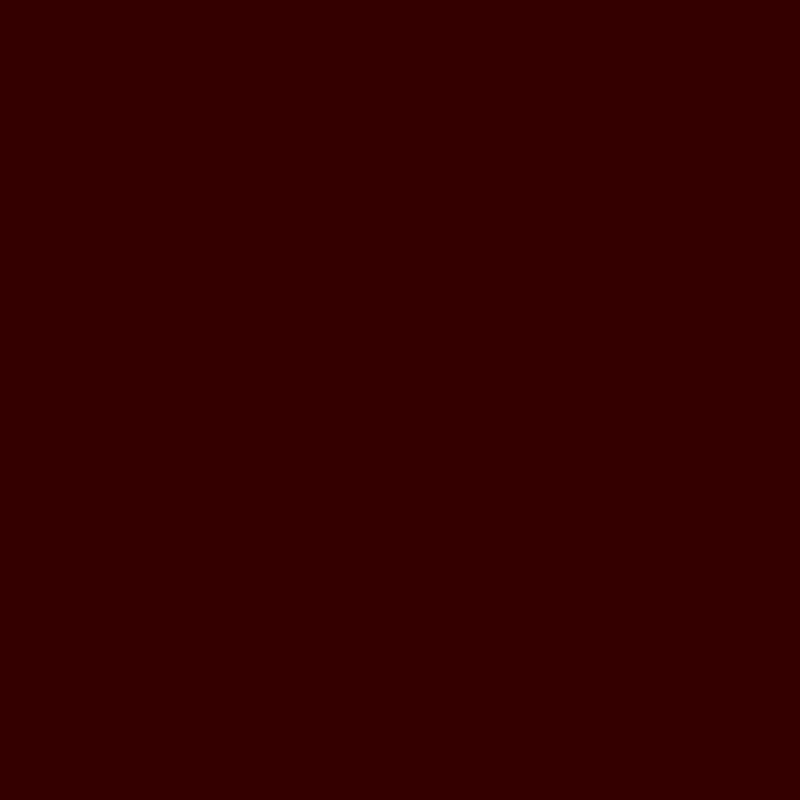}
        \caption{Color with night/low light}
        \label{fig:color_night_context}
    \end{subfigure}
    \hfill
    \begin{subfigure}[t]{0.23\textwidth}
        \centering
        \includegraphics[width=\textwidth]{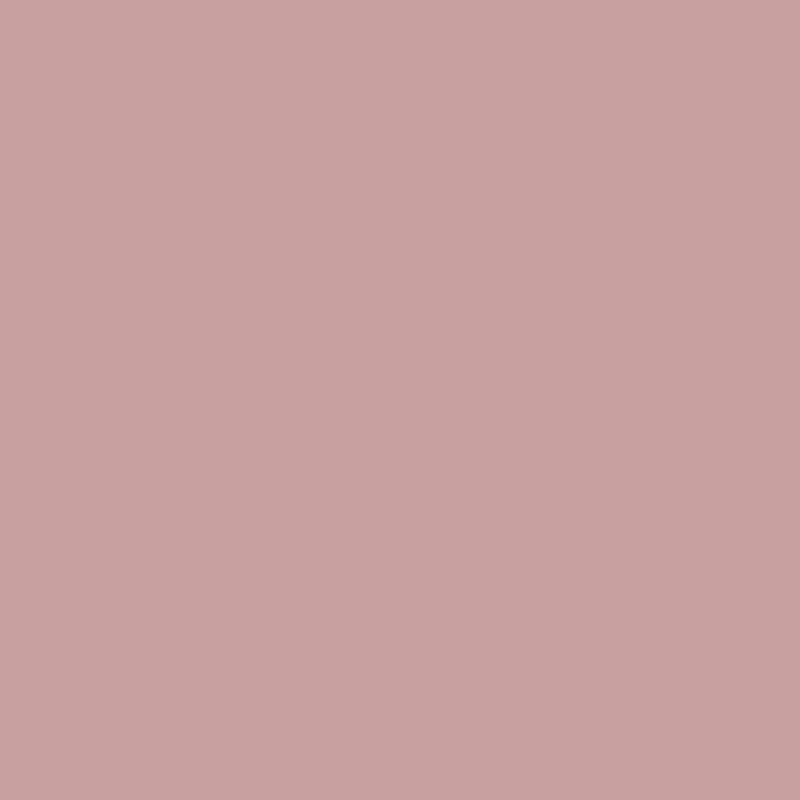}
        \caption{Color with fogginess}
        \label{fig:color_fog_context}
    \end{subfigure}
    \hfill
    \begin{subfigure}[t]{0.23\textwidth}
        \centering
        \includegraphics[width=\textwidth]{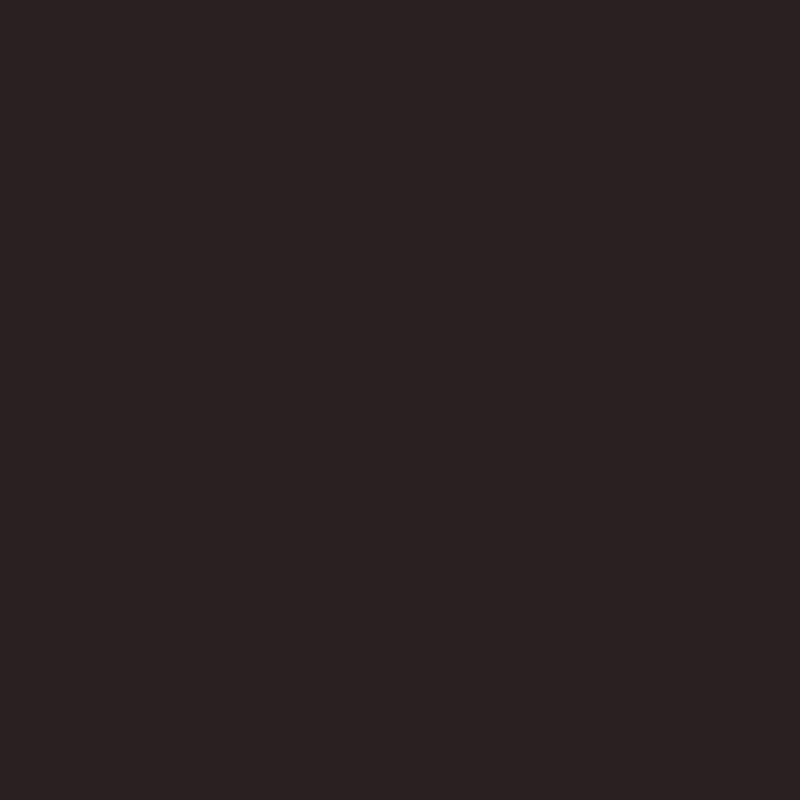}
        \caption{Color with night and fogginess}
        \label{fig:color_fog_night_context}
    \end{subfigure}
    \caption{Mean observed color for color-based error condition contexts.}
    \label{fig:color_weather_light_context}
\end{figure}

To utilize the above color similarity measure, we extract the RGB values for the region occupied by the agent. A key aspect is that we first include any contextual modifiers (e.g., weather/light) to account for changes to the color resulting from a given context. These contextual modifications can range from slight to severe. Additionally, the inclusion of context modifications demonstrates the sensitivity an agent might exhibit with respect to contextually based shifts in perceived color. To make our measurements, we extract the RGB values from the contextually modified observations. These extracted pixels are averaged and serve as the representative color for the entity (see Figure~\ref{fig:color_weather_light_context}). Given this representative color, the distance metric supports the failure likelihood estimate. The color distance indicates a possible, partial, or complete failure to detect an entity.

\subsubsection{Combined Contexts}\label{sec:combining_contexts}

As alluded to in the previous sections, it is possible to combine some/all the above state modifications to generate even more complex effects on the observations. For instance, Figure~\ref{fig:fog_night_context} demonstrates a case where the fog and lighting contexts are combined to generate an increased level of complexity and altered impact of each context parameter on the overall effect. Again, this is intended to further expand the scenarios we can represent. As with the individual contextual modifiers, the combined effects generate specific cases which can serve as analogs for states a driver might impact in a real driving setting. For example, it is reasonable to assume there are cases where humans or autonomous systems might encounter fog in darker driving conditions. By supporting these combined cases, we can see how changes in some parameters change the outcome for others. This is demonstrated in Figure~\ref{fig:color_weather_light_context}. As is apparent, the additional contextual details impact the perceived color of an entity. Consequently, this could impact the outcomes of color-sensitive cases.

An additional component of our combination of cases is how each driving agent is affected independent of the other. We do not assume that the two agents will be impacted equally by a particular context; instead, we allow for completely isolated assumptions regarding sensing for either agent. For instance, we could assume impacted sensing for the human driver based on some weather condition while assuming the autonomous sensing systems are unaffected. This configuration illustrates a significant disparity in the sensing capabilities in the given context, which we anticipate having a similarly significant impact on agent ability regarding detection and response.

\section{Simulation Modeling}\label{sec:simulation_modeling}

For our simulation and experimentation, our environment is based on the CARLO simulator introduced in \cite{cao2020reinforcement}, where it is used as one of the reference simulation environments. We chose this environment as it allows us to test our approach in a driving scenario while allowing us to restrict details of the driving task. CARLO reduces the environment to a two-dimensional representation of basic geometric shapes. This allows us to ignore extraneous details such as surface textures, complexity of object detection/recognition, etc. We can instead focus on a simulation of sensing and detection in this environment, and how changes in perception impact the driver performance in our hybrid team case.

\begin{figure}[ht]
	\centering
	\begin{subfigure}[t]{0.49\textwidth}
		\centering
		\includegraphics[width=\textwidth]{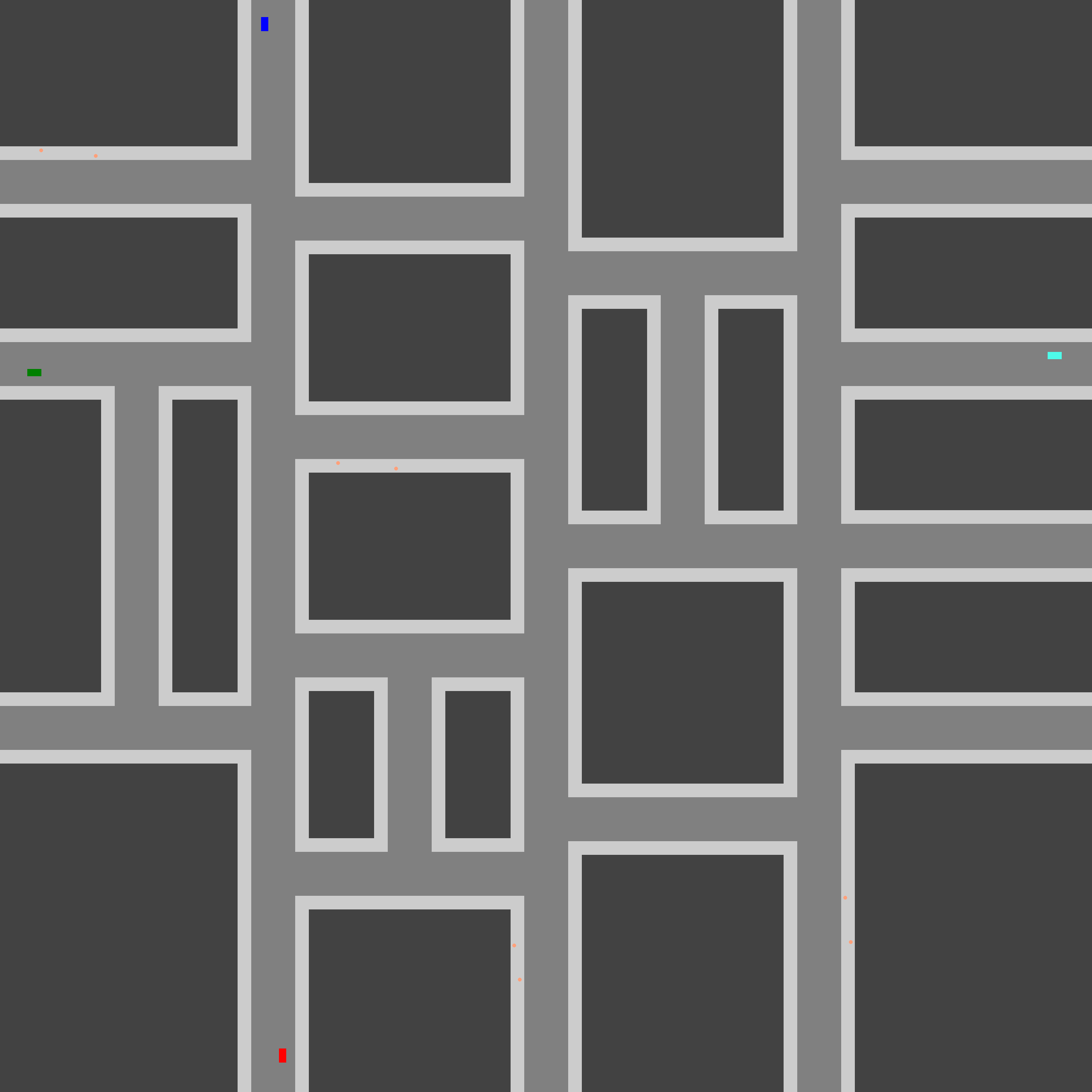}
		\caption{Main environment}
		\label{fig:sample_env}
	\end{subfigure}
	\hfill
	\begin{subfigure}[t]{0.49\textwidth}
		\centering
		\includegraphics[width=\textwidth]{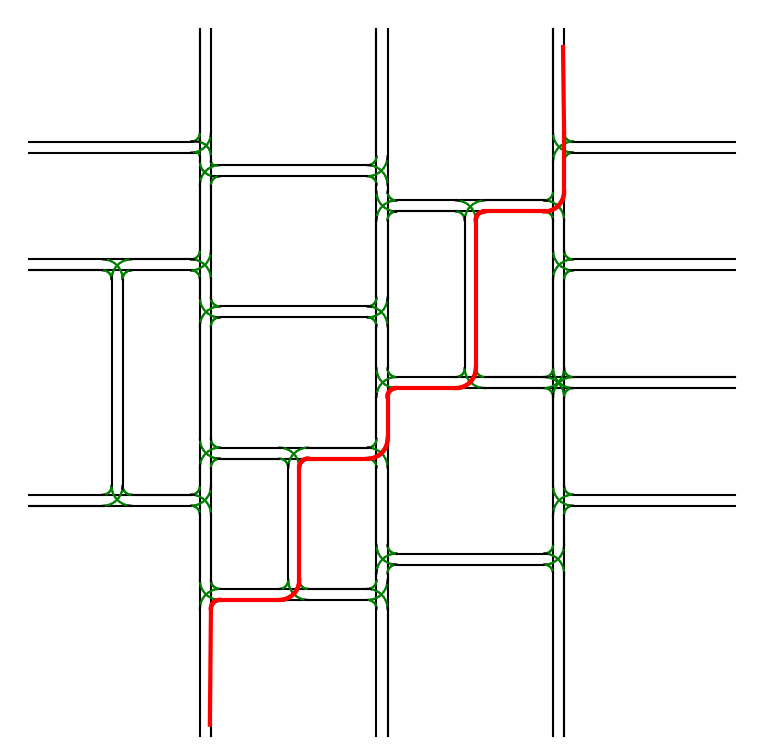}
		\caption{Sample path}
		\label{fig:sample_env_path}
	\end{subfigure}
	\caption{Sample Driving environment (including rectangular buildings, sidewalks, and cars) with shortest path trajectory}
	\label{fig:sample_env_and_path}
\end{figure}

The CARLO simulation environment provides a two-dimensional representation of a driving environment (see Figure~\ref{fig:sample_env}), which enables research in topics such as Automatic Control \cite{golnaraghi2017automatic}. The environment supports several entity types as well as graphical representation of scenes modeling an aerial view. The key supported entities include vehicles, pedestrians, and buildings. Additional entities can be used to define sidewalks and other basic shapes. In addition to the entities, CARLO enables basic control of movable objects (e.g., cars and pedestrians) via kinetic controls. Entities are initialized with positional and control parameters to define a starting state. Throughout the execution of an episode, the controls for acceleration and steering angle can be modified to generate turning behavior or changes in speed. With these controls, it is possible to simulate vehicles and pedestrians navigating the two-dimensional driving environment.

In addition to the entity generation and control, CARLO supports queries relating to inter-object distances and collision detection. The states of all entities in the environment are updated at a given discrete time step size (e.g., one update per $0.1s$). State updates generate updated positioning and kinematic values for each moving agent. Given these updated states, the queries provide a richer representation of the current world state. These are key to our purposes as we are then able to utilize parameters defining acceptable safe distances and related features. Further, collision detection allows for immediate awareness of an undesired outcome. With these, we can generate test cases which provide a basic, yet reasonable, analog for simplified driving cases.

\begin{figure}[ht]
	\centering
	\begin{subfigure}[t]{0.45\textwidth}
		\centering
		\includegraphics[width=\textwidth]{images/context_free.png}
		\caption{Four-way Intersection}
		\label{fig:sample_crossing_intersection}
	\end{subfigure}
	\hfill
	\begin{subfigure}[t]{0.45\textwidth}
		\centering
		\includegraphics[width=\textwidth]{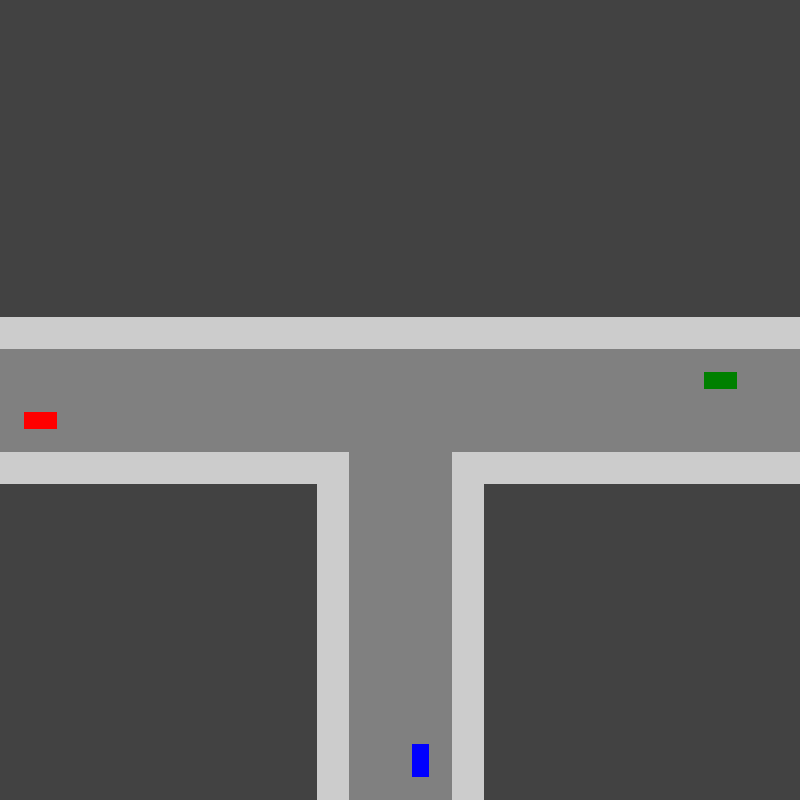}
		\caption{T-intersection}
		\label{fig:sample_t_intersection}
	\end{subfigure}
	\caption{Key interaction zones for driving scenario(s)}
\end{figure}

To demonstrate the effectiveness of our delegation manager, we chose to model several scenarios designed to recreate, or serve as an analog for, the cases as described in Subsection~\ref{sec:sensing_contexts}. Based on our simulation environment, where we use rectangular buildings and two-lane roads, we separate the problem space into two primary domains. First, the case where two roads intersect and generate a four-way intersection (see Figure~\ref{fig:sample_crossing_intersection}). This is of course a commonly observed road intersection, so we view it as a key case when demonstrating our approach. The second is the case of a road end intersecting a continuing road, which generates a T-Intersection (see Figure~\ref{fig:sample_t_intersection}). As with the four-way intersection, this is again a commonly encountered intersection type, so it is an additional and essential case for our scenarios. It is worth noting these two domains can be used to form the basis for interaction zones in a larger domain. As such, we will train and test our delegation manager in these domains to demonstrate how our approach can learn a desirable delegation model. Additionally, the use of these domains lets us draw specific attention to the key interaction zones. Further, we note that the use of a manager trained in these zones would support a more generalized use case (e.g., a Manhattan Scenario/Grid \cite{korger2008quality}).

\subsection{Scenarios}\label{sec:scenarios}

With our various supported contexts, to demonstrate the impact of context on performance, we generate scenarios including the above-mentioned sensing cases. These scenarios are parameterized to create two outcomes for human and autonomous drivers. More specifically, we identify parameters for contexts which can result in either successful or unsuccessful navigation of an environment when a driver operates independently. As a result, each combination of contexts generates a set of case-outcome pairs possible for a manager. In all but one, the manager can select at least one successful agent. The unavoidable failure case is simply the one in which both the human and autonomous drivers are unable to succeed in a context. In this case, the manager has no choice but to choose a failing driver for delegation. On the other hand, the remaining cases always include at least one driver the manager can delegate to achieve successful navigation of the scenario. 

To demonstrate the effectiveness of our delegation manager, we consider several measures of performance. First, we will show how the manager can learn a preference for drivers which can successfully navigate a given context. In other words, we will show that the manager learns to select an error-avoiding driver when possible. Next, we will compare the performance of the manager to that of a manager which simply selects drivers at random at each time step. The use of a random manager could of course still allow for successful navigation by chance, but at minimum would demonstrate a highly undesirable manager. More specifically, the random manager would in most cases be undesirable as it is likely to make too many frequent changes in the delegation. These rapid changes would make the driving task untenable as the human and autonomous system would be expected to rapidly gain and relinquish control. This would surely be exhausting and unsettling for a human driver, and this dynamic could increase the chances of failure during a change in control. To compare these various aspects of performance, we utilized the following scenario configurations in our experiments.

\subsubsection{Day Driving with Masks}\label{sec:day_driving_with_masks}

The daytime driving scenario serves to introduce errors with only minor changes made to the observation. In our experiments, we utilized a single mask in the forward sensing region. This enables error-induced and error-free behavior based on the severity level of the mask. For instance, a small mask at the farthest reaches of the sensing region would often have little or no impact on the driver's ability to detect other entities. However, increasing the mask's size in the forward region would result in a reduced ability to perceive entities in the direction the vehicle is traveling. As the size of the mask increases, the time between detection and error shrinks. For our experiments, we generate sensor and mask configurations which enable both success and failure cases for both human and autonomous drivers, illustrated in Table~\ref{tab:base_case_pairs}. Consequently, in each episode, the manager is observing one of four cases:

\begin{table}[H] 
    \caption{Human-AI Pair Cases. \label{tab:base_case_pairs}}
    \begin{tabularx}{\textwidth}{CCC}
        \toprule
        \textbf{Human Case}	& \textbf{Autonomous System Case} & \textbf{Expected Outcome}\\
        \midrule
        Error-free Mask & Error-free Mask & Success regardless of choice\\
        Error-free Mask & Error-inducing Mask & Success given human selected\\
        Error-inducing Mask & Error-free Mask & Success given AI selected\\
        Error-inducing Mask & Error-inducing Mask & Failure regardless of choice\\
        \bottomrule
    \end{tabularx}
\end{table}

\subsubsection{Day Driving with Fog}\label{sec:day_driving_with_fog}

For the fog scenario, errors are introduced through increasing severity of fog levels. In our experiments, we still utilize the sensing regions and masks, but only regions and masks which do not result in errors on their own. In other words, we only use sensing region and masks parameters from successful weather-free cases. To introduce errors, we utilize fog severity levels which generate failures in detection. For the error-free cases, we still include fog, but only at levels that do not degrade detection to the point of failure. A further key detail in this scenario is that we are only assuming a weather-based impact on sensing for the human driver. For the autonomous system, we are instead reusing the sensor and mask cases from the autonomous driver seen in Subsection~\ref{sec:day_driving_with_masks}. This demonstrates a case where humans are impacted by the weather conditions, but the autonomous system is instead only subject to their masks. This allows us to represent a disparity in the sensitivity to contexts between the human and autonomous system. As a result, the manager is learning to delegate in the following combinations of scenarios:

\begin{table}[H] 
    \caption{Human-AI Pair Cases. \label{tab:fog_case_pairs}}
    \begin{tabularx}{\textwidth}{CCC}
        \toprule
        \textbf{Human Case}	& \textbf{Autonomous System Case} & \textbf{Expected Outcome}\\
        \midrule
        Error-free Fog & Error-free Mask & Success regardless of choice\\
        Error-free Fog & Error-inducing Mask & Success given human selected\\
        Error-inducing Fog & Error-free Mask & Success given AI selected\\
        Error-inducing Fog & Error-inducing Mask & Failure regardless of choice\\
        \bottomrule
    \end{tabularx}
\end{table}

\subsubsection{Night Driving}\label{sec:night_driving}

Like the fog scenario, errors are introduced through increasing severity of lighting levels. As in the fog case, we are using differing levels of severity to either induce or omit erroneous behavior. These severity levels impact the darkness level, the width of the headlight region, and the distance covered by the headlight region. Further, the autonomous system is again not subject to the lighting conditions; instead, they will utilize the parameters seen in Subsection~\ref{sec:day_driving_with_masks}. Like the fog scenario, this demonstrates a case where the human is impacted by the lighting conditions, but the autonomous system is instead only subject to masks. This again allows us to represent a disparity in the sensitivity to contexts between the human and autonomous system. As a result, the manager is learning to delegate in the following combinations of scenarios:

\begin{table}[H] 
    \caption{Human-AI Pair Cases. \label{tab:light_case_pairs}}
    \begin{tabularx}{\textwidth}{CCC}
        \toprule
        \textbf{Human Case}	& \textbf{Autonomous System Case} & \textbf{Expected Outcome}\\
        \midrule
        Error-free Night & Error-free Mask & Success regardless of choice\\
        Error-free Night & Error-inducing Mask & Success given human selected\\
        Error-inducing Night & Error-free Mask & Success given AI selected\\
        Error-inducing Night & Error-inducing Mask & Failure regardless of choice\\
        \bottomrule
    \end{tabularx}
\end{table}

\subsubsection{Failed Color Detection}\label{sec:color_blindness_driving}

To represent failed detection or segmentation due to erroneous interpretation of sensing, we generate cases with an autonomous agent unable to perceive another vehicle in the environment due to its color. These cases will result in the autonomous system struggling to clearly recognize an entity exists. As such, this would lead to a failure to react to a potential upcoming collision. To offer success cases, we assign error colors which are not used for potential collision entities (i.e., unobserved or non-interacting entity color). Further, the human driver is not subject to the color-sensitivity conditions; instead, they will again utilize the parameters seen in Subsection~\ref{sec:day_driving_with_masks}. As a result, the manager is learning to delegate in the following combinations of scenarios:

\begin{table}[H] 
    \caption{Human-AI Pair Cases. \label{tab:color_case_pairs}}
    \begin{tabularx}{\textwidth}{CCC}
        \toprule
        \textbf{Human Case}	& \textbf{Autonomous System Case} & \textbf{Expected Outcome}\\
        \midrule
        Error-free Mask & Error-free Color & Success regardless of choice\\
        Error-free Mask & Error-inducing Color & Success given human selected\\
        Error-inducing Mask & Error-free Color & Success given AI selected\\
        Error-inducing Mask & Error-inducing Color & Failure regardless of choice\\
        \bottomrule
    \end{tabularx}
\end{table}

\subsubsection{Day Driving with Fog and AI Color Sensitivity}\label{sec:day_driving_with_ai_fog_and_color_blindness}

Extending on the previous fog case, the human driver’s success or failure is determined by the severity parameters. The distinction in this case is that we are no longer assuming the autonomous agent is unaffected by the fog. In this scenario, the assumption is that the autonomous system is sensitive to a particular color. This is further complicated by the fog levels, so we provide parameters which produce fog levels which modify the similarity of the observed color to the blindness color. This will of course demonstrate how the contextually based color can shift the perception toward or away from an error case. As a result, the manager is learning to delegate in the following combinations of scenarios:

\begin{table}[H] 
    \caption{Human-AI Pair Cases. \label{tab:fog_color_case_pairs}}
    \begin{tabularx}{\textwidth}{CCC}
        \toprule
        \textbf{Human Case}	& \textbf{Autonomous System Case} & \textbf{Expected Outcome}\\
        \midrule
        Error-free Fog & Error-free Fog-Color & Success regardless of choice\\
        Error-free Fog & Error-inducing Fog-Color & Success given human selected\\
        Error-inducing Fog & Error-free Fog-Color & Success given AI selected\\
        Error-inducing Fog & Error-inducing Fog-Color & Failure regardless of choice\\
        \bottomrule
    \end{tabularx}
\end{table}




\subsubsection{Noisy Estimates}\label{sec:noisy_observation_case}

As an additional test case, we wish to determine if adding noise to the generation of contexts impacts the manager’s performance. In this case, the observations provided to the manager are altered by some noise to introduce a divergence between the ground truth and the manager observation. The noise parameters affect the severity levels used to create the context-based observation. The degree to which the ground truth and observed state vary depend on the noise parameters. We sample new parameters via a truncated normal distribution centered at the current values. The truncation bounds and the standard deviation determine how widely the noisy estimate varies from the ground truth. To demonstrate these effects, we test the inclusion of noisy parameter estimates in the fog and night driving cases.

\section{Results}\label{sec:results}

In the following, we present and discuss the results for our trained delegation manager. The results represent manager performance in the various scenarios, including fog, night driving, masked sensors, etc. As noted in Subsection~\ref{sec:scenarios}, these contexts demonstrate cases where a human or autonomous driver could enter an undesired state (e.g., collision with another vehicle) as a result of a sensing/detection failure. We consider several aspects with respect to manager performance. First, we track the number of collisions the manager could avoid given the team dynamics. \emph{Avoidable} collisions are those which could have been avoided had the manager selected an error-free agent. For example, if the human's perception is impacted severely enough by fog to inhibit prompt detection of other vehicles, but the autonomous system is not, then a collision encountered when the human is delegated control would be considered avoidable. Hence, the manager should have detected the potential issue and selected the autonomous driver instead.

The next aspect we consider is the number of delegation changes. For our scenarios, we consider \emph{basic} delegation changes (i.e., a single and persistent change) and \emph{sudden} delegation changes. A sudden delegation change is a case where the manager changes delegation and then reverses the delegation in a sequence of three consecutive time steps (e.g., $\textrm{Human}\rightarrow\textrm{AI}\rightarrow\textrm{Human}$). As we noted previously, this is discouraged via a penalty in the manager's reward function. For basic delegation changes, we do not penalize the manager. Our intention was to focus on the more severe/inefficient case of sudden changes. It could be worth considering a penalty for total delegation choices in a modified reward function, but this was beyond the scope of our current work.

Given the reward function in Equation~\ref{eqn:manager_reward}, we can determine a sense of desirable manager behavior. In our scenarios, the episode lengths are larger than the positive reward received for a successful navigation outcome, so all episode rewards will be negative. The key distinction will of course be the magnitude of episode rewards. We expect a high-performing manager to identify successful driving agents and make few/no sudden delegation changes. Therefore, a high-performing manager should demonstrate similar performance values in the S/S, S/E, and E/S cases ($\textrm{E}\rightarrow\textrm{Error}, \textrm{S}\rightarrow\textrm{Success}$). Further, these episode rewards should be as close to zero as the episode lengths allow.

\subsection{Delegation Manager}\label{sec:results_delegation_manager}


The results presented in this section will demonstrate our manager's ability to identify necessary delegation decisions. Further, in all cases the results indicate the manager correctly identifies the suitable delegation(s) to avoid a collision with another vehicle. The performance of our manager is demonstrated in each of the scenarios we present. For each, the manager provides necessary delegations while still avoiding the discouraged cases of immediate delegation reversals.

As a general note regarding the following results, there are a few key aspects to remember. First, the episode lengths are used as a penalty, so the maximum achievable reward in our scenario is reduced. As noted in Equation~\ref{eqn:manager_reward}, the maximum reward for successful navigation is $100$ and the penalty for a collision is $-100$. After episode lengths, the remaining penalty is the number of immediate delegation reversals. With these factors, the expectation is that the manager performance across the two environments will be similar but will vary according to the episode length. In the case of the T-Intersection environment, the shortest expected episode is approximately $165$ time steps. For the Four-way Intersection environment, the shortest expected episode is slightly shorter at approximately $120$ time steps. Therefore, the best reward the manager could receive would be approximately $-65$ and $-25$ for the T-Intersection and Four-way Intersection environments, respectively.

Regarding the two delegation changes highlighted in the following results, we provide counts for two cases to illustrate related aspects of delegation. First, the basic changes reflect the manager's propensity to make delegation changes over an episode, but not cases where these changes are immediately reversed. The desire would still be a manager with fewer of these basic changes, but we do not explicitly discourage this behavior. Instead, our reward function focused on the sudden changes case. This case would be far more challenging in a realistic driving setting, so we prioritized it in our training and testing. As is demonstrated in our results, our reward function successfully discourages these immediate changes in delegation, preventing untenable manager behavior. As will be demonstrated in Section~\ref{sec:results_random_manager}, our use of the penalty significantly diminishes the occurrence of immediate delegation changes, especially when compared to a manager without this penalty.

Our manager's behavior and performance indicate both the ability to identify desirable delegations and reduce immediate delegation switches. For example, the results in Tables \ref{tab:h_normal_ai_normal} \& \ref{tab:h_weather_ai_normal} illustrate the approximately optimal path found by the team. Further, we see that each case included few or no immediate delegation changes. On the other hand, as expected, the scenarios where the manager has no choice but to select an erroneous agent show our manager is indeed encountering unavoidable failure cases. In both the T-Intersection and Four-way Intersection cases, we see the E/E cases resulting in significantly worse performance overall, which indicates our team is encountering the forced failure case. Therefore, the delegation has proven capable of associating the failure cases with desirable delegation choices. As seen in the following tables, the manager’s performance is consistently strong across our various scenarios. 

The results for our proposed manager are presented in Tables~\ref{tab:h_normal_ai_normal}-\ref{tab:h_noisy_weather_ai_normal}:

\begin{table}[H]
	\caption{Human: Sensor \& Mask, AI: Sensor \& Mask - test results for manager performance and behavior}
	\label{tab:h_normal_ai_normal}
	\begin{tabularx}{\textwidth}{CCCCC}
		\toprule
		\textbf{Human/AI Case} & \textbf{Avoidable Collisions} & \textbf{Mean Basic Changes} & \textbf{Mean Sudden Changes} & \textbf{Mean Reward}\\
		\midrule
		\multicolumn{5}{c}{\textbf{T-Intersection}}\\
		\midrule
		S/E & 0 & $1.0\pm 0.0$ & $0.0\pm 0.0$ & $-67.0\pm 0.0$\\
		E/S & 0 & $6.0\pm 0.0$ & $2.0\pm 0.0$ & $-69.0\pm 0.0$\\
		S/S & 0 & $6.1\pm 0.0$ & $0.1\pm 0.0$ & $-67.1\pm 0.1$\\
		E/E & 0 & $6.0\pm 0.0$ & $1.0\pm 0.0$ & $-184.0\pm 0.0$\\
		\midrule
		\multicolumn{5}{c}{\textbf{Four-way Intersection}}\\
		\midrule
		S/E & 0 & $3.0\pm 0.0$ & $0.0\pm 0.0$ & $-25.0\pm 0.0$\\
		E/S & 0 & $5.0\pm 0.0$ & $0.0\pm 0.0$ & $-20.0\pm 0.0$\\
		S/S & 0 & $1.0\pm 0.0$ & $0.0\pm 0.0$ & $-25.0\pm 0.0$\\
		E/E & 0 & $0.0\pm 0.0$ & $0.0\pm 0.0$ & $-184.0\pm 0.0$\\
		\bottomrule
	\end{tabularx}
\end{table}

\begin{table}[H]
	\caption{Human: Weather, AI: Sensor \& Mask - test results for manager performance and behavior}
	\label{tab:h_weather_ai_normal}
	\begin{tabularx}{\textwidth}{CCCCC}
		\toprule
		\textbf{Human/AI Case} & \textbf{Avoidable Collisions} & \textbf{Mean Basic Changes} & \textbf{Mean Sudden Changes} & \textbf{Mean Reward}\\
		\midrule
		\multicolumn{5}{c}{\textbf{T-Intersection}}\\
		\midrule
		S/E & 0 & $10.2\pm 0.0$ & $1.6\pm 0.0$ & $-69.0\pm 0.1$\\
		E/S & 0 & $10.8\pm 0.0$ & $3.4\pm 0.0$ & $-65.6\pm 0.5$\\
		S/S & 0 & $4.6\pm 0.0$ & $0.3\pm 0.0$ & $-68.0\pm 0.0$\\
		E/E & 0 & $6.7\pm 0.0$ & $0.9\pm 0.0$ & $-144.6\pm 16.5$\\
		\midrule
		\multicolumn{5}{c}{\textbf{Four-way Intersection}}\\
		\midrule
		S/E & 0 & $2.0\pm 0.0$ & $0.0\pm 0.0$ & $-22.0\pm 0.0$\\
		E/S & 0 & $0.0\pm 0.0$ & $0.0\pm 0.0$ & $-25.0\pm 0.0$\\
		S/S & 0 & $0.0\pm 0.0$ & $0.0\pm 0.0$ & $-25.0\pm 0.0$\\
		E/E & 0 & $0.0\pm 0.0$ & $0.0\pm 0.0$ & $-180.7\pm 6.3$\\
		\bottomrule
	\end{tabularx}
\end{table}

\begin{table}[H]
	\caption{Human: Light, AI: Sensor \& Mask - test results for manager performance and behavior}
	\label{tab:h_light_ai_normal}
	\begin{tabularx}{\textwidth}{CCCCC}
		\toprule
		\textbf{Human/AI Case} & \textbf{Avoidable Collisions} & \textbf{Mean Basic Changes} & \textbf{Mean Sudden Changes} & \textbf{Mean Reward}\\
		\midrule
		\multicolumn{5}{c}{\textbf{T-Intersection}}\\
		\midrule
		S/E & 0 & $14.0\pm 0.0$ & $3.0\pm 0.0$ & $-70.0\pm 0.0$\\
		E/S & 0 & $17.9\pm 0.0$ & $2.4\pm 0.0$ & $-69.8\pm 0.3$\\
		S/S & 0 & $22.0\pm 0.0$ & $2.0\pm 0.0$ & $-69.0\pm 0.0$\\
		E/E & 0 & $11.1\pm 0.0$ & $2.1\pm 0.0$ & $-186.3\pm 0.2$\\
		\midrule
		\multicolumn{5}{c}{\textbf{Four-way Intersection}}\\
		\midrule
		S/E & 0 & $9.0\pm 0.0$ & $1.0\pm 0.0$ & $-26.0\pm 0.1$\\
		E/S & 0 & $8.0\pm 0.0$ & $3.1\pm 0.0$ & $-28.1\pm 0.1$\\
		S/S & 0 & $20.0\pm 0.0$ & $4.0\pm 0.0$ & $-29.0\pm 0.0$\\
		E/E & 0 & $6.4\pm 0.0$ & $0.0\pm 0.0$ & $-118.4\pm 22.3$\\
		\bottomrule
	\end{tabularx}
\end{table}

\begin{table}[H]
	\caption{Human: Sensor \& Mask, AI: Color - test results for manager performance and behavior}
	\label{tab:h_normal_ai_color}
	\begin{tabularx}{\textwidth}{CCCCC}
		\toprule
		\textbf{Human/AI Case} & \textbf{Avoidable Collisions} & \textbf{Mean Basic Changes} & \textbf{Mean Sudden Changes} & \textbf{Mean Reward}\\
		\midrule
		\multicolumn{5}{c}{\textbf{T-Intersection}}\\
		\midrule
		S/E & 0 & $4.1\pm 0.0$ & $0.0\pm 0.0$ & $-67.0\pm 0.1$\\
		E/S & 0 & $12.1\pm 0.0$ & $3.1\pm 0.0$ & $-74.3\pm 0.1$\\
		S/S & 0 & $7.1\pm 0.0$ & $2.0\pm 0.0$ & $-69.0\pm 0.1$\\
		E/E & 0 & $10.0\pm 0.0$ & $0.0\pm 0.0$ & $-185.0\pm 0.0$\\
		\midrule
		\multicolumn{5}{c}{\textbf{Four-way Intersection}}\\
		\midrule
		S/E & 0 & $6.0\pm 0.0$ & $0.0\pm 0.0$ & $-27.0\pm 0.1$\\
		E/S & 0 & $1.2\pm 0.0$ & $0.0\pm 0.0$ & $-25.0\pm 0.0$\\
		S/S & 0 & $2.0\pm 0.0$ & $0.4\pm 0.0$ & $-25.4\pm 0.1$\\
		E/E & 0 & $2.0\pm 0.0$ & $0.0\pm 0.0$ & $-184.0\pm 0.0$\\
		\bottomrule
	\end{tabularx}
\end{table}


\begin{table}[H]
	\caption{Human: Weather, AI: Weather \& Color - test results for manager performance and behavior}
	\label{tab:h_weather_ai_weather_color}
	\begin{tabularx}{\textwidth}{CCCCC}
		\toprule
		\textbf{Human/AI Case} & \textbf{Avoidable Collisions} & \textbf{Mean Basic Changes} & \textbf{Mean Sudden Changes} & \textbf{Mean Reward}\\
		\midrule
		\multicolumn{5}{c}{\textbf{T-Intersection}}\\
		\midrule
		S/E & 0 & $21.5\pm 0.0$ & $8.6\pm 0.0$ & $-77.8\pm 3.2$\\
		E/S & 0 & $25.4\pm 0.0$ & $11.9\pm 0.0$ & $-93.6\pm 13.9$\\
		S/S & 0 & $30.6\pm 0.0$ & $16.5\pm 0.0$ & $-93.4\pm 11.8$\\
		E/E & 0 & $5.6\pm 0.0$ & $1.9\pm 0.0$ & $-162.4\pm 13.7$\\
		\midrule
		\multicolumn{5}{c}{\textbf{Four-way Intersection}}\\
		\midrule
		S/E & 0 & $1.0\pm 0.0$ & $0.0\pm 0.0$ & $-25.0\pm 0.0$\\
		E/S & 0 & $0.0\pm 0.0$ & $0.0\pm 0.0$ & $-25.0\pm 0.0$\\
		S/S & 0 & $0.0\pm 0.0$ & $0.0\pm 0.0$ & $-25.0\pm 0.0$\\
		E/E & 0 & $8.0\pm 0.0$ & $3.0\pm 0.0$ & $-173.9\pm 12.3$\\
		\bottomrule
	\end{tabularx}
\end{table}

\begin{table}[H]
	\caption{Human: Noisy Light, AI: Sensor \& Mask - test results for manager performance and behavior}
	\label{tab:h_noisy_light_ai_normal}
	\begin{tabularx}{\textwidth}{CCCCC}
		\toprule
		\textbf{Human/AI Case} & \textbf{Avoidable Collisions} & \textbf{Mean Basic Changes} & \textbf{Mean Sudden Changes} & \textbf{Mean Reward}\\
		\midrule
		\multicolumn{5}{c}{\textbf{T-Intersection}}\\
		\midrule
		S/E & 0 & $7.1\pm 0.0$ & $2.7\pm 0.0$ & $-70.1\pm 0.3$\\
		E/S & 0 & $10.5\pm 0.0$ & $3.3\pm 0.0$ & $-70.3\pm 0.5$\\
		S/S & 0 & $6.2\pm 0.0$ & $1.1\pm 0.0$ & $-68.1\pm 0.2$\\
		E/E & 0 & $0.1\pm 0.0$ & $0.0\pm 0.0$ & $-184.4\pm 0.1$\\
		\midrule
		\multicolumn{5}{c}{\textbf{Four-way Intersection}}\\
		\midrule
		S/E & 0 & $3.2\pm 0.0$ & $0.1\pm 0.0$ & $-25.1\pm 0.1$\\
		E/S & 0 & $16.0\pm 0.0$ & $5.2\pm 0.0$ & $-25.2\pm 0.6$\\
		S/S & 0 & $6.7\pm 0.0$ & $1.0\pm 0.0$ & $-26.0\pm 0.3$\\
		E/E & 0 & $7.9\pm 0.0$ & $2.2\pm 0.0$ & $-146.6\pm 19.4$\\
		\bottomrule
	\end{tabularx}
\end{table}

\begin{table}[H]
	\caption{Human: Noisy Weather, AI: Sensor \& Mask - test results for manager performance and behavior}
	\label{tab:h_noisy_weather_ai_normal}
	\begin{tabularx}{\textwidth}{CCCCC}
		\toprule
		\textbf{Human/AI Case} & \textbf{Avoidable Collisions} & \textbf{Mean Basic Changes} & \textbf{Mean Sudden Changes} & \textbf{Mean Reward}\\
		\midrule
		\multicolumn{5}{c}{\textbf{T-Intersection}}\\
		\midrule
		S/E & 0 & $13.9\pm 0.0$ & $5.2\pm 0.0$ & $-72.7\pm 0.5$\\
		E/S & 0 & $11.2\pm 0.0$ & $4.1\pm 0.0$ & $-71.1\pm 0.6$\\
		S/S & 0 & $7.8\pm 0.0$ & $2.7\pm 0.0$ & $-69.7\pm 0.3$\\
		E/E & 0 & $6.1\pm 0.0$ & $1.9\pm 0.0$ & $-96.9\pm 15.6$\\
		\midrule
		\multicolumn{5}{c}{\textbf{Four-way Intersection}}\\
		\midrule
		S/E & 0 & $5.0\pm 0.0$ & $2.0\pm 0.0$ & $-27.0\pm 0.0$\\
		E/S & 0 & $2.4\pm 0.0$ & $1.2\pm 0.0$ & $-26.2\pm 0.1$\\
		S/S & 0 & $12.4\pm 0.0$ & $3.8\pm 0.0$ & $-28.8\pm 0.3$\\
		E/E & 0 & $4.0\pm 0.0$ & $1.6\pm 0.0$ & $-185.6\pm 0.2$\\
		\bottomrule
	\end{tabularx}
\end{table}

As the above results indicate, our delegation manager successfully identifies desirable agents for delegation while reducing the occurrence of undesirable immediate delegation changes. Further, we see consistently strong results across the various contexts presented. The primary exception would be the ``Noisy Weather, AI: Sensor \& Mask'' case. In this case, we see improved performance in the E/E case. This is can likely be attributed to the fact that our error cases only occur if the agents reach the interaction point at the same time. If the agent observations are skewed enough by the context, it is feasible for the conflicting vehicles to reach the interaction point at an interval which enables accidental avoidance. This, combined with a manager operating on noisy observations, could alter the behavior of the managed driving agents, which could reduce the likelihood of reaching the given failure scenario. Therefore, we believe this occurred in this case, resulting in slightly improved (suboptimal) performance. It is still apparent that the results indicate a remaining significant increase in failures, so the manager's performance is still demonstrated.
 
As an additional side effect of the noisy observations, the degradation of the observations based on the weather or lighting conditions will obscure more of distinguishing features of a given state, which could lead to higher uncertainty in the manager's decisions. Further, this could shift the manager's preference between either driver toward uniform. In this case, the manager would have an approximately equal preference between the two agents, so we would expect an increased occurrence of basic changes as indicated in our results. In future approaches, we could investigate altered reward functions to discourage other frequencies of delegation changes. We could either penalize any delegation change or vary the magnitude of the penalty by type of delegation change. As indicated by the results, these changes would likely offer little change regarding the success of navigation but would reduce the likelihood of changes in manager delegations.

From the results demonstrated, we conclude that our manager has learned to both identify the agents likely to lead to successful navigation and avoid immediate delegation changes. This is accomplished with a reward function which offers little in the way of domain knowledge. The manager is not provided rewards specific to a driving task, only details regarding how successful the team performed, and how well the manager avoids undesirable delegation behavior. Therefore, we conclude the proposed manager model can learn a suitable association between contexts, behavior, and desirable decisions.

\section{Random Manager}\label{sec:results_random_manager}

We determined that a purely random manager with no limit on delegation interval would have an unfair advantage over the learned manager in our setting. This is because the random manager would not respect a penalty discouraging frequent immediate delegation changes. As a result, the random manager can make delegation choices frequently enough that it could by chance perform a sequence of delegations and complete a driving episode. This would be beneficial if the ability to make multiple delegation changes in rapid fashion were not a problem for the user. To counteract this, we introduced the interval which blocks the delegation decision until the end of the interval (e.g., once every $10$ time steps). This was intended to illustrate two key factors. First, how sensitive the scenario is to the frequency of delegation decisions. Second, how much of a gap in performance is seen between the interval-restricted manager and learning manager in the same contexts. When comparing our delegation manager to the random manager, the effect of numerous immediate switches would still be apparent in agent reward, but we believed this would a less significant representation of performance overall. As a final note, we will not be testing the noisy scenarios for the random manager as the noise will not impact the manager's behavior. Therefore, the results for the noisy case are already indicated by the noise-free cases.

The results for the random manager are presented in Tables~\ref{tab:rand_h_normal_ai_normal}-\ref{tab:rand_h_weather_ai_weather_color}:


\begin{table}[H]
	\caption{Mean rewards and avoidable collision counts (by Human-AI configuration) for Random Manager with given delegation interval lengths in the Human: Sensor \& Mask, AI: Sensor \& Mask case.}
	\label{tab:rand_h_normal_ai_normal}
	\begin{tabularx}{\textwidth}{CCCCCCCC}
		\toprule
		\textbf{Human/AI} & \textbf{10} & \textbf{15} & \textbf{20} & \textbf{25} & \textbf{30} & \textbf{35} & \textbf{40}\\
		\midrule
		\multicolumn{8}{c}{\textbf{T-Intersection}}\\
		\midrule
		S/E & $-97.3\pm 19.3$ & $-108.3\pm 15.9$ & $-101.7\pm 18.6$ & $-142.3\pm 25.2$ & $-117.9\pm 16.5$ & $-130.1\pm 16.7$ & $-114.5\pm 22.5$\\
		E/S & $-87.0\pm 12.9$ & $-92.9\pm 13.9$ & $-127.0\pm 17.2$ & $-124.9\pm 22.7$ & $-124.2\pm 16.5$ & $-136.1\pm 17.0$ & $-140.8\pm 19.4$\\
		S/S & $-67.0\pm 0.0$ & $-67.0\pm 0.0$ & $-67.0\pm 0.0$ & $-67.0\pm 0.0$ & $-67.0\pm 0.0$ & $-67.0\pm 0.0$ & $-67.0\pm 0.0$\\
		E/E & $-186.8\pm 0.7$ & $-186.0\pm 0.8$ & $-186.3\pm 0.8$ & $-185.8\pm 0.8$ & $-186.0\pm 0.8$ & $-185.7\pm 0.8$ & $-186.0\pm 0.8$\\
		\midrule
		\multicolumn{8}{c}{\textbf{Avoidable Collisions}}\\
		\midrule
		S/E & 4 & 17 & 10 & 10 & 21 & 26 & 8\\
		E/S & 9 & 11 & 23 & 10 & 24 & 26 & 24\\
		\midrule
		\multicolumn{8}{c}{\textbf{Four-way Intersection}}\\
		\midrule
		S/E & $-33.5\pm 10.5$ & $-82.0\pm 21.2$ & $-53.4\pm 17.0$ & $-81.1\pm 21.4$ & $-66.0\pm 19.4$ & $-58.7\pm 18.5$ & $-79.1\pm 20.9$\\
		E/S & $-26.9\pm 6.2$ & $-56.5\pm 17.7$ & $-78.8\pm 20.9$ & $-71.7\pm 20.4$ & $-72.7\pm 20.2$ & $-68.7\pm 19.9$ & $-113.0\pm 22.6$\\
		S/S & $-25.0\pm 0.0$ & $-25.0\pm 0.0$ & $-25.0\pm 0.0$ & $-25.0\pm 0.0$ & $-25.0\pm 0.0$ & $-25.0\pm 0.0$ & $-25.0\pm 0.0$\\
		E/E & $-184.0\pm 0.0$ & $-184.0\pm 0.0$ & $-184.0\pm 0.0$ & $-184.0\pm 0.0$ & $-184.0\pm 0.0$ & $-184.0\pm 0.0$ & $-184.0\pm 0.0$\\
		\midrule
		\multicolumn{8}{c}{\textbf{Avoidable Collisions}}\\
		\midrule
		S/E & 3 & 18 & 9 & 18 & 13 & 11 & 17\\
		E/S & 1 & 10 & 17 & 15 & 15 & 14 & 26\\
		\bottomrule
	\end{tabularx}
    \bigskip
    \caption{Mean rewards and avoidable collision counts (by Human-AI configuration) for Random Manager with given delegation interval lengths in the Human: Weather, AI: Sensor \& Mask case.}
    \label{tab:rand_h_weather_ai_normal}
    \begin{tabularx}{\textwidth}{CCCCCCCC}
    	\toprule
    	\textbf{Human/AI} & \textbf{10} & \textbf{15} & \textbf{20} & \textbf{25} & \textbf{30} & \textbf{35} & \textbf{40}\\
    	\midrule
    	\multicolumn{8}{c}{\textbf{T-Intersection}}\\
    	\midrule
    	S/E & $-88.8\pm 15.5$ & $-98.3\pm 15.6$ & $-103.4\pm 17.6$ & $-152.6\pm 24.4$ & $-130.1\pm 16.7$ & $-115.8\pm 16.4$ & $-119.3\pm 19.7$\\
    	E/S & $-119.4\pm 20.8$ & $-97.4\pm 17.6$ & $-134.5\pm 21.4$ & $-139.9\pm 23.7$ & $-123.0\pm 16.3$ & $-135.6\pm 16.7$ & $-133.2\pm 17.3$\\
    	S/S & $-67.5\pm 0.1$ & $-67.3\pm 0.1$ & $-67.3\pm 0.1$ & $-67.4\pm 0.1$ & $-71.7\pm 8.3$ & $-67.4\pm 0.1$ & $-67.3\pm 0.1$\\
    	E/E & $-181.3\pm 9.1$ & $-162.1\pm 17.2$ & $-173.1\pm 14.2$ & $-182.0\pm 9.1$ & $-181.5\pm 9.1$ & $-186.6\pm 0.6$ & $-186.8\pm 0.6$\\
    	\midrule
    	\multicolumn{8}{c}{\textbf{Avoidable Collisions}}\\
    	\midrule
    	S/E & 6 & 11 & 12 & 14 & 26 & 20 & 17\\
    	E/S & 12 & 8 & 17 & 12 & 22 & 26 & 26\\
    	\midrule
    	\multicolumn{8}{c}{\textbf{Four-way Intersection}}\\
    	\midrule
    	S/E & $-23.7\pm 0.4$ & $-66.1\pm 19.4$ & $-53.3\pm 17.0$ & $-75.2\pm 20.7$ & $-78.9\pm 20.9$ & $-71.7\pm 20.4$ & $-111.1\pm 22.4$\\
    	E/S & $-36.6\pm 12.1$ & $-53.3\pm 17.0$ & $-66.0\pm 19.4$ & $-52.4\pm 17.1$ & $-91.8\pm 21.8$ & $-55.7\pm 17.8$ & $-98.1\pm 22.0$\\
    	S/S & $-25.0\pm 0.0$ & $-25.0\pm 0.0$ & $-25.0\pm 0.0$ & $-25.0\pm 0.0$ & $-25.0\pm 0.0$ & $-25.0\pm 0.0$ & $-25.0\pm 0.0$\\
    	E/E & $-184.0\pm 0.0$ & $-184.0\pm 0.0$ & $-184.0\pm 0.0$ & $-184.0\pm 0.0$ & $-184.0\pm 0.0$ & $-184.0\pm 0.0$ & $-184.0\pm 0.0$\\
    	\midrule
    	\multicolumn{8}{c}{\textbf{Avoidable Collisions}}\\
    	\midrule
    	S/E & 0 & 13 & 9 & 16 & 17 & 15 & 26\\
    	E/S & 4 & 9 & 13 & 9 & 21 & 10 & 23\\
    	\bottomrule
    \end{tabularx}
\end{table}
\begin{table}[H]
    \caption{Mean rewards and avoidable collision counts (by Human-AI configuration) for Random Manager with given delegation interval lengths in the Human: Light, AI: Sensor \& Mask case.}
    \label{tab:rand_h_night_ai_normal}
    \begin{tabularx}{\textwidth}{CCCCCCCC}
        \toprule
        \textbf{Human/AI} & \textbf{10} & \textbf{15} & \textbf{20} & \textbf{25} & \textbf{30} & \textbf{35} & \textbf{40}\\
        \midrule
        \multicolumn{8}{c}{\textbf{T-Intersection}}\\
        \midrule
        S/E & $-116.3\pm 21.5$ & $-111.0\pm 16.0$ & $-97.1\pm 15.0$ & $-169.4\pm 27.6$ & $-131.3\pm 16.9$ & $-135.8\pm 17.2$ & $-117.4\pm 19.5$\\
        E/S & $-91.1\pm 14.9$ & $-101.8\pm 17.0$ & $-126.2\pm 19.4$ & $-124.2\pm 23.4$ & $-104.5\pm 15.1$ & $-121.1\pm 16.2$ & $-170.9\pm 20.4$\\
        S/S & $-67.0\pm 0.0$ & $-67.0\pm 0.0$ & $-67.0\pm 0.0$ & $-67.0\pm 0.0$ & $-67.0\pm 0.0$ & $-67.0\pm 0.0$ & $-67.0\pm 0.0$\\
        E/E & $-186.1\pm 0.7$ & $-185.3\pm 0.7$ & $-181.6\pm 9.2$ & $-185.3\pm 0.7$ & $-176.8\pm 12.1$ & $-185.5\pm 0.7$ & $-186.3\pm 0.7$\\
        \midrule
        \multicolumn{8}{c}{\textbf{Avoidable Collisions}}\\
        \midrule
        S/E & 10 & 18 & 13 & 7 & 26 & 26 & 16\\
        E/S & 9 & 11 & 20 & 8 & 15 & 23 & 26\\
        \midrule
        \multicolumn{8}{c}{\textbf{Four-way Intersection}}\\
        \midrule
        S/E & $-43.2\pm 14.4$ & $-40.4\pm 13.3$ & $-59.6\pm 18.3$ & $-85.1\pm 21.5$ & $-69.1\pm 19.9$ & $-52.3\pm 17.1$ & $-88.6\pm 21.6$\\
        E/S & $-27.0\pm 6.2$ & $-40.5\pm 13.3$ & $-85.1\pm 21.5$ & $-58.6\pm 18.5$ & $-69.1\pm 19.9$ & $-61.6\pm 19.1$ & $-109.3\pm 22.2$\\
        S/S & $-25.0\pm 0.0$ & $-25.0\pm 0.0$ & $-25.0\pm 0.0$ & $-25.0\pm 0.0$ & $-25.0\pm 0.0$ & $-25.0\pm 0.0$ & $-25.0\pm 0.0$\\
        E/E & $-177.9\pm 11.7$ & $-141.8\pm 23.7$ & $-184.0\pm 0.0$ & $-177.9\pm 11.7$ & $-184.0\pm 0.0$ & $-177.9\pm 11.7$ & $-177.9\pm 11.7$\\
        \midrule
        \multicolumn{8}{c}{\textbf{Avoidable Collisions}}\\
        \midrule
        S/E & 6 & 5 & 11 & 19 & 14 & 9 & 20\\
        E/S & 1 & 5 & 19 & 11 & 14 & 12 & 26\\
        \bottomrule
    \end{tabularx}
    \bigskip
    \caption{Mean reward (by Human-AI configuration) for Random Manager with given delegation interval lengths in the Human: Sensor \& Mask, AI: Color case.}
    \label{tab:rand_h_normal_ai_color}
    \begin{tabularx}{\textwidth}{CCCCCCCC}
        \toprule
        \textbf{Human/AI} & \textbf{10} & \textbf{15} & \textbf{20} & \textbf{25} & \textbf{30} & \textbf{35} & \textbf{40}\\
        \midrule
        \multicolumn{8}{c}{\textbf{T-Intersection}}\\
        \midrule
        S/E & $-80.5\pm 13.2$ & $-92.7\pm 13.2$ & $-109.3\pm 18.6$ & $-108.5\pm 18.7$ & $-128.3\pm 16.5$ & $-134.6\pm 16.7$ & $-121.4\pm 21.6$\\
        E/S & $-97.8\pm 14.5$ & $-126.6\pm 16.2$ & $-131.4\pm 16.3$ & $-130.7\pm 16.6$ & $-124.3\pm 16.1$ & $-121.9\pm 15.9$ & $-115.6\pm 15.5$\\
        S/S & $-70.4\pm 1.0$ & $-69.6\pm 0.8$ & $-79.5\pm 11.5$ & $-73.9\pm 1.6$ & $-70.4\pm 0.8$ & $-70.8\pm 0.9$ & $-99.8\pm 20.4$\\
        E/E & $-185.6\pm 0.6$ & $-185.6\pm 0.6$ & $-185.6\pm 0.6$ & $-185.8\pm 0.6$ & $-185.5\pm 0.6$ & $-185.3\pm 0.6$ & $-185.4\pm 0.6$\\
        \midrule
        \multicolumn{8}{c}{\textbf{Avoidable Collisions}}\\
        \midrule
        S/E & 2 & 10 & 12 & 9 & 25 & 26 & 12\\
        E/S & 13 & 25 & 26 & 26 & 24 & 23 & 20\\
        \midrule
        \multicolumn{8}{c}{\textbf{Four-way Intersection}}\\
        \midrule
        S/E & $-26.1\pm 1.1$ & $-89.7\pm 21.3$ & $-67.5\pm 18.2$ & $-85.7\pm 21.4$ & $-57.8\pm 17.5$ & $-50.2\pm 16.2$ & $-90.0\pm 21.3$\\
        E/S & $-33.8\pm 10.5$ & $-75.8\pm 20.6$ & $-50.2\pm 16.2$ & $-65.2\pm 19.5$ & $-53.8\pm 16.9$ & $-68.8\pm 19.9$ & $-121.1\pm 23.2$\\
        S/S & $-25.0\pm 0.0$ & $-25.0\pm 0.0$ & $-25.0\pm 0.0$ & $-25.0\pm 0.0$ & $-25.0\pm 0.0$ & $-25.0\pm 0.0$ & $-25.0\pm 0.0$\\
        E/E & $-184.0\pm 0.0$ & $-184.0\pm 0.0$ & $-184.0\pm 0.0$ & $-184.0\pm 0.0$ & $-184.0\pm 0.0$ & $-184.0\pm 0.0$ & $-184.0\pm 0.0$\\
        \midrule
        \multicolumn{8}{c}{\textbf{Avoidable Collisions}}\\
        \midrule
        S/E & 0 & 20 & 12 & 19 & 10 & 8 & 20\\
        E/S & 3 & 16 & 8 & 13 & 9 & 14 & 26\\
        \bottomrule
    \end{tabularx}
\end{table}

\begin{table}[H]
	\caption{Mean rewards and avoidable collision counts (by Human-AI configuration) for Random Manager with given delegation interval lengths in the Human: Weather, AI: Weather \& Color case.}
	\label{tab:rand_h_weather_ai_weather_color}
	\begin{tabularx}{\textwidth}{CCCCCCCC}
		\toprule
		\textbf{Human/AI} & \textbf{10} & \textbf{15} & \textbf{20} & \textbf{25} & \textbf{30} & \textbf{35} & \textbf{40}\\
		\midrule
		\multicolumn{8}{c}{\textbf{T-Intersection}}\\
		\midrule
		S/E & $-97.7\pm 18.4$ & $-94.6\pm 18.2$ & $-99.2\pm 17.1$ & $-125.3\pm 24.2$ & $-85.5\pm 13.0$ & $-132.6\pm 18.2$ & $-139.2\pm 24.2$\\
		E/S & $-103.3\pm 18.5$ & $-89.4\pm 13.4$ & $-106.7\pm 17.9$ & $-125.9\pm 20.2$ & $-109.6\pm 18.0$ & $-126.8\pm 18.0$ & $-141.6\pm 18.0$\\
		S/S & $-77.8\pm 12.0$ & $-77.7\pm 12.1$ & $-71.7\pm 8.3$ & $-68.9\pm 3.2$ & $-71.8\pm 8.3$ & $-74.2\pm 9.4$ & $-92.7\pm 17.0$\\
		E/E & $-172.6\pm 14.2$ & $-172.6\pm 14.1$ & $-176.8\pm 11.9$ & $-165.0\pm 16.4$ & $-176.6\pm 12.0$ & $-165.4\pm 16.3$ & $-158.7\pm 17.5$\\
		\midrule
		\multicolumn{8}{c}{\textbf{Avoidable Collisions}}\\
		\midrule
		S/E & 5 & 5 & 9 & 6 & 8 & 23 & 12\\
		E/S & 10 & 10 & 13 & 18 & 15 & 21 & 26\\
		\midrule
		\multicolumn{8}{c}{\textbf{Four-way Intersection}}\\
		\midrule
		S/E & $-30.1\pm 8.7$ & $-34.2\pm 10.5$ & $-37.4\pm 12.0$ & $-48.8\pm 16.4$ & $-43.8\pm 14.3$ & $-68.9\pm 19.9$ & $-112.7\pm 22.7$\\
		E/S & $-42.5\pm 14.5$ & $-59.3\pm 18.4$ & $-62.5\pm 18.9$ & $-75.0\pm 20.7$ & $-56.0\pm 17.7$ & $-65.2\pm 19.5$ & $-131.1\pm 23.5$\\
		S/S & $-25.0\pm 0.0$ & $-25.0\pm 0.0$ & $-25.0\pm 0.0$ & $-25.0\pm 0.0$ & $-25.0\pm 0.0$ & $-25.0\pm 0.0$ & $-25.0\pm 0.0$\\
		E/E & $-177.9\pm 11.7$ & $-184.0\pm 0.0$ & $-184.0\pm 0.0$ & $-167.0\pm 18.2$ & $-177.9\pm 11.7$ & $-167.0\pm 18.2$ & $-184.0\pm 0.0$\\
		\midrule
		\multicolumn{8}{c}{\textbf{Avoidable Collisions}}\\
		\midrule
		S/E & 2 & 3 & 4 & 8 & 6 & 14 & 26\\
		E/S & 6 & 11 & 12 & 16 & 10 & 13 & 26\\
		\bottomrule
	\end{tabularx}
\end{table}

As indicated in the results, in many cases there is a bit of variance across the number of avoidable collisions as the interval length increases. For instance, Table~\ref{tab:rand_h_normal_ai_normal} demonstrates possible periodicity in the performance. We see smaller numbers of avoidable collisions in some intervals (e.g., $4/9$ for T-Intersection with $10$ step interval) while seeing higher numbers in others. We attribute this to several aspects of the scenarios. First, the fact that the manager is behaving completely randomly can by chance still allow for randomly successful outcomes. Second, the time step where the detection is essential versus the interval length is important. If the interval divides almost equally (e.g., $85$ steps in the T-Intersection case), then the random manager has a better chance of randomly selecting the correct agent in time. The correct agent in the key time step triggers a response for collision avoidance, so an interval which aligns well with these key states could improve team chances of success. Further, this key time step is different between the two environments, so we can see these values vary between the two environments. A final factor would be the fact that the longer interval lengths reduce the number of overall delegation decisions made. Therefore, the likelihood of a correct choice at the right time can increase with this compatibility between interval lengths and key time steps.

Despite the observed variance in performance with respect to avoidable collisions, the results still clearly indicate the improvements made from a learned delegation manager policy over a random manager. Each case demonstrates how the manager both chooses to make significantly more sudden delegation changes and makes far fewer desirable delegations. In all the cases where the manager can fail (i.e., S/E, E/S, and E/E), we see far more cases of avoidable collisions and significantly smaller rewards. Further, we see a much more severe variance in manager performance versus the learned manager. This again shows how the random manager can select correct delegations, but the rate of success and consistency of performance are much lower than the learned model. For example, with respect to the performance of the learned manager (see Table~\ref{tab:h_normal_ai_normal}), we note a drop in performance up to $127\%$ and $390\%$ in rewards for the T-Intersection and Four-way Intersection cases, respectively. Additionally, compared to zero for the learned manager, we see an increase to up to $26$ cases of avoidable collisions. Combined, these aspects indicate the delegation manager's significance with respect to team performance. A manager failing to properly utilize the proper agent clearly fails to avoid collisions. This further indicates that the use of a trained manager enables improved performance beyond that of cases where either agent would be allowed to operate without intervention.

\section{Conclusion}\label{sec:conclusion}

\subsection{Final Remarks}\label{sec:final_remarks}

As our results indicate, the inclusion of a learned delegation behavior policy improves team performance. The improvement in performance is a result of the manager identifying a desirable driver and corresponding delegation of control. We further emphasize the impact of our manager by comparison with a random manager. We acknowledge that a random manager with the ability to make frequent delegation changes could by chance select the appropriate driver. On the other hand, we also note how the random manager's use of frequent delegation changes would make it incompatible with a real setting. A human user would most likely be unable or unwilling to use a delegation manager that is so unpredictable. The user would be burdened by the random manager's unpredictable behavior and subsequently be burdened with a higher cognitive load. Therefore, we demonstrate an effective and delegation-efficient manager.

\subsection{Future Work}\label{sec:future_work}

As we acknowledged earlier, our simulator and test environments are a simplification of realistic driving cases. In one aspect, we perform a simplification through reduced fidelity in the simulation. Our use of basic geometric shapes and kinematics reduces the alignment to real-world driving. As a future approach, we could investigate extending this approach to simulators with higher fidelity in their models, such as the CARLA simulation environment \cite{Dosovitskiy17} or DeepDrive \cite{craig_quiter_2018_1248998}. Another simplification we utilize in our scenarios is the reduced size of our driving scenarios. We utilized our environments to specifically highlight the consequences of sensing deficiencies in highly instructive cases. While this enables our demonstration of performance in key interaction cases, given the strong manager performance, we could investigate extended driving scenarios with more complex paths and more potential cases for interaction. These changes would demonstrate how well our method translates to a more true-to-life setting, but we cannot say with certainty whether it would provide an improved demonstration of the delegation learning itself.

An additional aspect we could investigate, to see how well our results translate to new cases, would be the inclusion of more managed vehicles in a single environment. With the additional vehicles, and a larger map, the tested vehicles would demonstrate longer driving sequences and multiple interactions. Further, the inclusion of additional managed vehicles would indicate how well two or more managed vehicles can account for each other. On the other hand, it is our belief this demonstrates less a case of compensating for sensing failures and more a case of improved driving capabilities. Similarly, within our CARLO scenario, we could further extend our environments to include more complexity in roads and related. For instance, we could add traffic signals at intersections to modify the failure scenarios to include cases involving missed traffic signals. In these cases, we could simulate an agent failing to detect a traffic signal, resulting in the agent entering the intersection at the wrong time. These scenarios would align well with the Computer Vision concepts we noted in Section~\ref{sec:related_work}. They could demonstrate cases where sensors either fail to detect an object or a perturbation was made to prevent detection.

\acknowledgments{This work was partly supported by: the H2020 Humane-AI-Net project (grant No. 952026), CHIST-ERA (grant No. CHIST-ERA-19-XAI-010-MUR-484-22), and by the European Union under the Italian National Recovery and Resilience Plan (NRRP) of partnership on “Artificial Intelligence: Foundational Aspects” (PE0000013 - program “FAIR”).}

\begin{adjustwidth}{-\extralength}{0cm}

\reftitle{References}

\bibliography{referenced_papers.bib}

\PublishersNote{}
\end{adjustwidth}

\end{document}